\pdfoutput=1
\documentclass[conference,compsoc]{IEEEtran}

\usepackage{xfp}

\usepackage{tikz}
\usepackage{amsmath}
\usepackage[utf8]{inputenc}
\usepackage{siunitx}

\usepackage[most]{tcolorbox}
\usepackage{xcolor}

\usepackage{enumitem}
\usepackage{tikz}
\usetikzlibrary{calc}
\usepackage{tikz}
\usetikzlibrary{calc, positioning}
\usepackage{pifont}
\usepackage{caption}
\usepackage{booktabs}
\usepackage{multirow}
\definecolor{mygreen}{RGB}{0,205,0}
\usepackage{hyperref}
\usepackage[ruled,vlined,linesnumbered]{algorithm2e}
\usepackage{amsmath, amssymb}
\usepackage{colortbl}
\usepackage{threeparttable}
\usepackage{makecell}

\usepackage{array}

\newcommand{\colorASRA}[1]{%
  \ifnum#1>90
    \cellcolor{green!15}#1
  \else\ifnum#1>75
    \cellcolor{green!10}#1
  \else\ifnum#1>69
    \cellcolor{yellow!10}#1
  \else\ifnum#1>59
    \cellcolor{orange!10}#1
  \else
    \cellcolor{red!10}#1
  \fi\fi\fi\fi
}

\newcommand{\colorASRB}[1]{%
  \begingroup
  \ifdim #1pt > 80pt
    \cellcolor{green!25}#1%
  \else\ifdim #1pt > 75pt
    \cellcolor{green!20}#1%
  \else\ifdim #1pt > 70pt
    \cellcolor{green!15}#1%
  \else\ifdim #1pt > 60pt
    \cellcolor{green!10}#1%
  \else\ifdim #1pt > 50pt
    \cellcolor{yellow!10}#1%
  \else\ifdim #1pt > 40pt
    \cellcolor{orange!10}#1%
  \else
    \cellcolor{red!10}#1%
  \fi\fi\fi\fi\fi\fi
  \endgroup
}

\newcommand{\colorGCS}[1]{%
  \ifdim #1pt > 90pt
    \cellcolor{green!15}#1
  \else\ifdim #1pt > 83pt
    \cellcolor{green!10}#1
  \else\ifdim #1pt > 75pt
    \cellcolor{yellow!10}#1
  \else
    \cellcolor{orange!10}#1
  \fi\fi\fi
}

\usepackage{xspace}
\usepackage{graphicx}
\usetikzlibrary{shapes.misc,arrows.meta,fit,backgrounds}
\newcommand{\sys}{\textsc{Snoop}\xspace}
\newcommand{\nsd}{SOR\xspace}

\definecolor{atkred}{RGB}{225,30,30}
\definecolor{defgreen}{RGB}{30,132,73}
\definecolor{dropgreen}{RGB}{0,175,55}
\definecolor{snoopblue}{RGB}{47,85,151}
\definecolor{lightbg}{RGB}{245,247,250}
\definecolor{cCodex}{HTML}{1F5F8B}
\definecolor{cOpen}{HTML}{E59A4C}
\definecolor{cIter}{HTML}{7FB3D5}
\definecolor{cODR}{HTML}{9C7A54}
\newcommand{\aCodex}{Codex}
\newcommand{\aOpen}{OpenHands}
\newcommand{\aIter}{IterResearch}
\newcommand{\aODR}{ODR}
\usepackage[most]{tcolorbox}
\newtcolorbox{findingbox}{enhanced,colback=snoopblue!5,colframe=snoopblue!55,
  boxrule=0.6pt,arc=2pt,left=5pt,right=5pt,top=2.5pt,bottom=2.5pt,boxsep=1pt,
  before skip=5pt,after skip=5pt}
\newcommand{\finding}[3][]{\begin{findingbox}\textbf{Finding-#2#1:}\ #3\end{findingbox}}

\newcommand{\red}[1]{\textcolor{atkred}{#1}}
\newcommand{\vd}[2]{\ensuremath{#1_{\textcolor{dropgreen}{\downarrow#2}}}}
\providecommand{\vu}[2]{\ensuremath{#1_{\textcolor{atkred}{\uparrow#2}}}}

\usepackage{tikz}

\newcommand{\circnum}[1]{%
  \tikz[baseline=(char.base)]{
    \node[shape=circle,fill=black,
          text=white,font=\bfseries\scriptsize,
          inner sep=1pt] (char) {#1};
  }%
}

\usepackage{pifont}

\begin{document}
\date{}
\title{Actions Speak Louder than Words: Stateful Privacy Sniffing of\\ LLM Agents via Behavioral Covert Channels in Web Interactions}

\author{
\IEEEauthorblockN{Ruichao Liang\IEEEauthorrefmark{1},
Wenzhong Deng\IEEEauthorrefmark{2},
Jing Chen\IEEEauthorrefmark{2},
Yuqing Li\IEEEauthorrefmark{2},
Kaixuan Li\IEEEauthorrefmark{1},
Cong Wu\IEEEauthorrefmark{2},
Yang Liu\IEEEauthorrefmark{1}}
\IEEEauthorblockA{\IEEEauthorrefmark{1}Nanyang Technological University\\
\{ruichao.liang, kaixuan.li, yangliu\}@ntu.edu.sg}
\IEEEauthorblockA{\IEEEauthorrefmark{2}Wuhan University\\
\{2024302181016, chenjing, li.yuqing, cnacwu\}@whu.edu.cn}
}

\maketitle

\begin{abstract}
LLM agents increasingly rely on Web interaction, issuing searches, following links, and reading pages, to complete complex user tasks. Since each action is driven by the agent's private context, including the user's prompt, dialogue history, memory, and connected data, it also opens a new privacy exposure surface. Prior attacks either passively analyze network traffic, which yields only coarse correlations, or hijack the agent through indirect prompt injection, which must place sensitive content in an outgoing field and is blocked by content-level defenses. Whether an attacker can actively recover fine-grained private context without any explicit transmission thus remains insufficiently understood.

Our insight is that an agent's benign Web decisions themselves leak private context, as what it chooses to visit encodes \emph{whether, which, and in what order} it acts, forming a \emph{behavioral covert channel}. Building on this, we design \sys, a stateful hierarchical probing framework that narrows a coarse-to-fine hypothesis tree per task and reconstructs user queries and profiles from the agent's observable paths. Across four agents and three LLM backbones on 500 StackExchange queries and 378 multi-turn iKAT dialogues, \sys recovers query concepts with an average F1 of 0.89 and user profiles with an F1 of 0.80, while four representative agent defenses change its recovery by less than 0.03 F1. Since no sensitive content is ever transmitted, content-level inspection cannot detect this leakage; we therefore propose \emph{Selection-Obfuscated Retrieval} (\nsd), which acquires resources under a fixed budget independent of private context and confines the agent's selection to a local snapshot, cutting \sys's F1 to 0.39 without additional LLM inference.
\end{abstract}

\section{Introduction}

Large language models (LLMs) have evolved into autonomous agents capable of executing complex tasks~\cite{yao2023react,schick2023toolformer}.
By integrating external tools, these agents can interact with their environments, gather information, and perform multi-step tasks.
Among these capabilities, Web access is one of the most widely adopted, allowing agents to search for up-to-date information, retrieve online resources, and adapt their actions based on newly acquired information.
However, Web interactions may expose sensitive user information, as agents have access to users' private context, including query prompts, conversation histories, and task memory, raising significant privacy concerns.

\begin{figure}[t]
\centering
\includegraphics[width=0.47\textwidth]{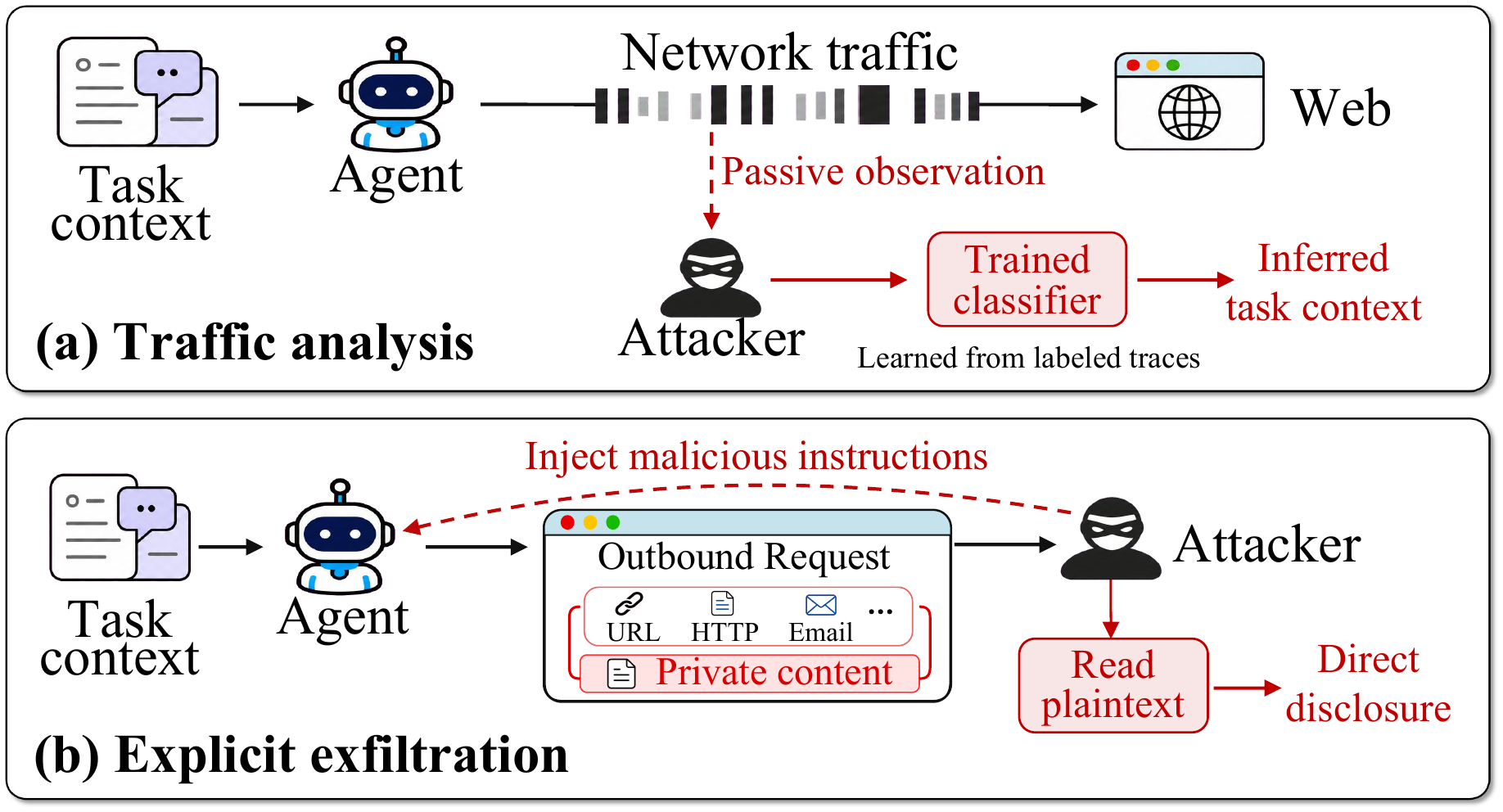}
\caption{Existing approaches to agent privacy leakage.}
\label{fig:Privacy}
\end{figure}

Prior studies have investigated privacy leakage in LLM agents through two primary approaches.
\circnum{1} Some investigate privacy leakage through \textit{traffic analysis} (Figure~\ref{fig:Privacy}a).
Jeong et al.~\cite{jeong2026wra} analyze the domain sequences generated by agents accessing external websites, while AgentPrint~\cite{zhang2025agentprint} examines encrypted traffic between users and cloud-based agents.
SkillWatermark~\cite{li2026skillwatermark} further demonstrates privacy inference through network traffic patterns, including packet sizes and timing.
\circnum{2} Others investigate \textit{explicit privacy exfiltration} (Figure~\ref{fig:Privacy}b), where attackers inject malicious instructions to induce agents to transmit private information to attacker-controlled destinations~\cite{greshake2023not,wu2024wipi}.
Such exfiltration typically embeds sensitive content in URL parameters, HTTP request bodies, emails, or other outbound communications~\cite{liao2025eia,aimlabs2025echoleak}.

However, existing approaches have limitations that leave privacy risks in LLM agents insufficiently explored.
Firstly, privacy inference based on traffic analysis primarily relies on passively observed network traces and learned correlations between traffic patterns and private information, limiting targeted, fine-grained privacy reconstruction.
Secondly, explicit exfiltration attacks require inducing agents to disclose private information through outgoing requests or tool outputs.
Although techniques such as encoding and obfuscation can conceal sensitive content, malicious instructions, explicit exfiltration attempts, and suspiciously encoded payloads still remain detectable by model-level safety mechanisms and content-based defenses.
These limitations raise a fundamental question: \emph{Can attackers actively probe an agent's private context at a fine-grained level without explicitly exfiltrating sensitive information?}

\textbf{Insights}.
Our key observation is that LLM agents make Web access decisions during task execution, such as which webpages to visit, links to follow, and resources to retrieve.
These decisions are influenced by the agent's private context, while the resulting requests are observable to external websites.
By presenting semantically distinct resources, an attacker could induce the agent to make context-dependent choices and infer private information from the selected destinations.
We refer to this mechanism as a \emph{behavioral covert channel}, in which private information is encoded in observable Web actions.
\circnum{1} Unlike traffic-based privacy inference, which primarily relies on passive classification of coarse-grained traffic patterns, this channel enables active, fine-grained probing of private information.
\circnum{2} Unlike explicit exfiltration attacks, it offers greater stealth by avoiding direct transmission of sensitive content, making it less susceptible to model-level safeguards and content-based defenses.
\circnum{3} Moreover, unlike existing approaches, it requires no privileged network observer (e.g., ISP, VPN provider, DNS resolver)~\cite{jeong2026wra,zhang2025agentprint,li2026skillwatermark}, relying only on requests to attacker-controlled endpoints.
To systematically characterize this privacy threat, we investigate four research questions:

\begin{itemize}[leftmargin=1.1em,itemsep=1.5pt,topsep=2pt]
    \item \textbf{RQ1:} How do existing LLM agents perform Web interactions, and can their interaction workflows expose private user information through observable behavior?
    \item \textbf{RQ2:} What are the capacity and reliability of different behavioral covert channels?
    \item \textbf{RQ3:} How can isolated behavioral leaks be leveraged for continuous, fine-grained privacy reconstruction?
    \item \textbf{RQ4:} To what extent can existing agent defenses mitigate privacy leakage through behavioral covert channels?
\end{itemize}

To investigate these questions, we conduct an empirical study of Web interaction workflows across four representative LLM agents and three LLM backbones.
By introducing conditional content into retrieved webpages and observing the resulting Web requests, we identify three key findings.
\textbf{Finding-1:} LLM agents' Web interactions follow an \emph{observe--reason--act} loop, where retrieved Web content re-enters the agent's reasoning process and, together with private user context, influence subsequent Web access decisions.
\textbf{Finding-2:} Six behavioral covert channels exhibit varying capacities and reliability. Choice-based channels provide the most effective trade-off, while other channels are more susceptible to variations in agent execution and network conditions.
\textbf{Finding-3:} Isolated behavioral probes are insufficient for continuous, fine-grained privacy reconstruction. Effective recovery requires hierarchical, adaptive, and stateful probing to progressively refine privacy hypotheses and handle intermittent Web interactions.

\textbf{\sys.}
Guided by these findings, we propose \sys, a stateful privacy-probing framework that systematically exploits behavioral covert channels in LLM agents' Web interactions to probe private user information. It enables continuous, fine-grained privacy reconstruction through budget-constrained, adaptive probing over a hierarchical hypothesis space.
Specifically, \sys (i) maintains a \textit{Snoop Tree} to organize and progressively refine privacy hypotheses for each search task; (ii) employs an \textit{Adaptive Probe Constructor} to select probing directions and generate semantically distinct candidates; (iii) uses a \textit{Snoop Agenda} to coordinate probing across tasks, preserve state, and adapt to unsuccessful probes; and (iv) performs \textit{Privacy Reconstruction} to aggregate behavioral evidence into reconstructed user queries and profiles.
Overall, \sys enables stealthy, active, and fine-grained privacy probing without explicit data exfiltration or privileged network observation.

We conduct comprehensive evaluations of \sys across four representative, widely adopted LLM agents and three state-of-the-art (SOTA) LLM backbones.
We use two complementary benchmarks: 500 StackExchange questions~\cite{stackexchange} spanning eight domains and 27 topics for query-level reconstruction, and 378 conversational turns from 26 personas in TREC iKAT~\cite{aliannejadi2024ikat} for multi-turn profiling.
Our evaluation covers 240 distinct configurations and 63,216 query-level execution opportunities, yielding 55,351 behavioral-path records, over 412,000 probing-step records, and 2,159 profile-level records.
\sys achieves an average query-level concept F1 of 0.89 (up to 0.97) and profile F1 of 0.80, outperforming existing traffic-based inference and explicit exfiltration attacks.
It requires only 6.5 and 5.9 probes per query on StackExchange and iKAT, respectively, while retaining 0.87 answer coverage with a median answer-length increase of 1.3$\times$.
We further deploy four SOTA agent defenses: Spotlighting~\cite{hines2024spotlighting}, Firewalls~\cite{sharma2025firewalls}, Task Shield~\cite{jia2025taskshield}, and Fides~\cite{costa2025fides}, none of which reduces reconstruction F1 by more than 0.03.
These results reveal that seemingly legitimate Web interactions can expose private user information while largely evading existing defenses.

\begin{figure*}[t]
\centering
\includegraphics[width=0.92\textwidth]{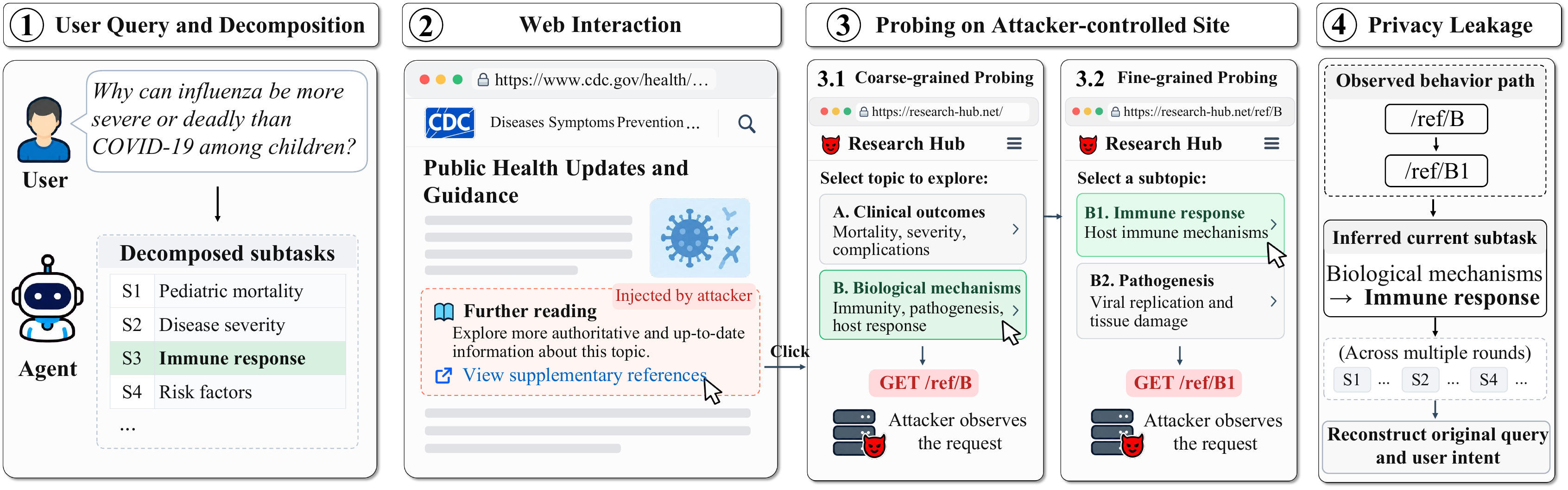}
\caption{Motivating example of behavioral privacy leakage.}
\label{fig:motivating}
\end{figure*}

\textbf{Mitigation.}
To mitigate behavioral privacy leakage, we propose \emph{Selection-Obfuscated Retrieval} (\nsd), which reduces the correlation between observable Web access patterns and private user context.
The key idea is to decouple external Web access from private-context-dependent agent reasoning through context-independent acquisition, local semantic selection, and randomized retrieval.
Experiments show that \nsd reduces attack reconstruction F1 from 0.89 to 0.39 at the query level and from 0.80 to 0.39 at the profile level, without additional LLM inference.
Overall, this paper makes the following contributions:
\begin{itemize}[leftmargin=1.1em,itemsep=1.5pt,topsep=2pt]
  \item We reveal a previously underexplored privacy threat in LLM agents: private user information can leak through context-dependent Web interactions without explicit data exfiltration. We systematically characterize behavioral covert channels, highlighting externally observable agent decisions as a distinct privacy attack surface.
  \item We develop \sys, a stateful privacy-probing framework that exploits behavioral covert channels for stealthy, active, and fine-grained privacy inference. It adaptively explores hierarchical privacy hypotheses to reconstruct user queries and profiles without explicit data exfiltration or privileged network observation.
  \item We evaluate \sys across diverse agents, models, and privacy-reconstruction tasks, demonstrating effective and efficient privacy recovery while exposing critical gaps in existing agent defenses against behavioral leakage.
  \item We propose \emph{Selection-Obfuscated Retrieval} (\nsd), a behavioral privacy defense that decouples external Web access from private-context-dependent reasoning, substantially reducing attack reconstruction accuracy by obscuring private semantic choices.

\end{itemize}

\section{Background and Related Work}

\subsection{Agent Web Interactions and Execution Context}

LLM agents interact with Web to search for information, retrieve resources, and accomplish complex tasks~\cite{yao2023react,zhou2024webarena,deng2023mind2web}.
Representative systems include browser automation agents (e.g., Browser-Use~\cite{browseruse2025}), research agents (e.g., WebResearcher and IterResearch~\cite{qiao2025webresearcher,chen2025iterresearch}), and general-purpose agents (e.g., Codex and OpenHands~\cite{openai2025codex,wang2025openhands}).
Despite architectural differences, these agents incorporate retrieved Web content into their reasoning to determine subsequent search and browsing actions.
Such decisions are informed by two sources of context: \emph{private context}, including user queries, conversation history, and task memory; and \emph{external context}, consisting of information retrieved from the Web.
Their interaction shapes the agent's subsequent Web access decisions, which may be observable to external websites.
We investigate two levels of private information: \emph{query-level information}, capturing the intent and concepts of individual user queries; and \emph{profile-level information}, capturing user attributes, interests, and preferences accumulated across multiple interactions.

\subsection{Privacy Leakage and Defenses in LLM Agents}

Prior studies have demonstrated privacy leakage through side-channel signals during LLM inference, including response lengths, timing, and packet sizes~\cite{weiss2024prompt,carlini2024timing,mcdonald2025whisperleak}. In LLM agents, WRA~\cite{jeong2026wra} infers private information from visited domain sequences, while AgentPrint~\cite{zhang2025agentprint} analyzes encrypted traffic between users and cloud-based agents. SkillWatermark~\cite{li2026skillwatermark} further exploits traffic patterns induced by conditional instructions in agent skills.
Another line of research injects malicious instructions for explicit privacy exfiltration, inducing agents to transmit sensitive information to attacker-controlled destinations through URLs, tool arguments, or other outbound requests~\cite{greshake2023not,wu2024wipi,liao2025eia,fu2024imprompter,aimlabs2025echoleak}. Benchmarks such as AgentDojo, InjecAgent, ASB, and WASP further assess agents' susceptibility to such injection attacks, including data exfiltration~\cite{debenedetti2024agentdojo,zhan2024injecagent,zhang2025asb,evtimov2025wasp}.

Existing agent defenses mitigate malicious instructions and unauthorized information disclosure through different mechanisms. Content-level defenses distinguish trusted instructions from untrusted content or filter suspicious inputs and outputs~\cite{hines2024spotlighting,chen2025struq,wallace2024hierarchy,sharma2025firewalls,liu2025datasentinel}.
Task-level defenses restrict actions inconsistent with user objectives~\cite{jia2025taskshield,zhu2025melon}, while information-flow controls track and constrain sensitive data propagation~\cite{costa2025fides,debenedetti2025camel,shi2025progent,wu2025isolategpt}.
However, these defenses largely overlook behavioral covert channels, where an agent's legitimate actions in Web interactions reveal private context without explicitly transmitting sensitive data or violating task constraints.
Such leakage is difficult to detect through content filtering, task-alignment checks, or information-flow controls.

\section{Motivating Example}
\label{sec:background}

As a simplified example, Figure~\ref{fig:motivating} illustrates how an attacker can progressively infer private task information through behavioral probing.
Consider a research agent answering the query ``\emph{Why can influenza be more severe or deadly than COVID-19 among children?}''
The agent decomposes the query into subtasks concerning pediatric mortality, disease severity, immune response, and risk factors.
While investigating immune response, it visits a medical webpage, where an attacker-provided supplementary reference link redirects it to an attacker-controlled probing site.

The attacker then conducts two rounds of probing through legitimate reference selections
The first presents \emph{Clinical Outcomes} and \emph{Biological Mechanisms}, leading the agent to select the latter and issue \texttt{GET /ref/B}.
The second distinguishes \emph{Immune Response} from \emph{Pathogenesis}, yielding \texttt{GET /ref/B1}.
From observable requests, the attacker infers the private subtask \emph{Biological Mechanisms $\rightarrow$ Immune Response}, without accessing the original query or receiving sensitive content.
By aggregating probing paths across additional subtasks and interactions, the attacker can further reconstruct the original query's semantics and intent.
This illustrates how successive behavioral choices enable progressively finer-grained privacy reconstruction.

\section{Threat Model}\label{sec:threat}

\subsection{System Setting}

We consider LLM agents that autonomously plan and execute multi-step tasks in open environments, with access to external websites and online services.
The system consists of three entities: the user who specifies the task, the LLM agent that executes it, and the external Web environment with which the agent interacts as needed.

\subsection{Attacker Capabilities}

We assume an attacker with the following capabilities and constraints.
(i) The attacker \textbf{CAN} host a probing website and monitor incoming requests, including accessed endpoints, request counts, ordering, timing, etc.
(ii) The attacker \textbf{CAN} inject a supplementary reference link or pop-up into a webpage encountered by the agent during normal task execution, redirecting it to the attacker-controlled probing website.
(iii) The attacker has \textbf{NO} prior knowledge of the user's original query or the injected page's topic or content, and \textbf{CANNOT} observe traffic to other websites.
(iv) The attacker \textbf{CANNOT} modify the agent or its underlying model, or observe network traffic through privileged infrastructure such as an ISP, VPN provider, or DNS resolver.

\subsection{Attack Goals}

We consider two privacy reconstruction goals.
(i) Query-level reconstruction aims to recover the intent and concepts of an individual user query from behavioral observations and synthesize a semantically equivalent query.
(ii) Profile-level reconstruction aims to infer user attributes, interests, and preferences by aggregating recovered information across multiple queries and interactions.

\begin{figure}[t]
\centering
\includegraphics[width=0.48\textwidth]{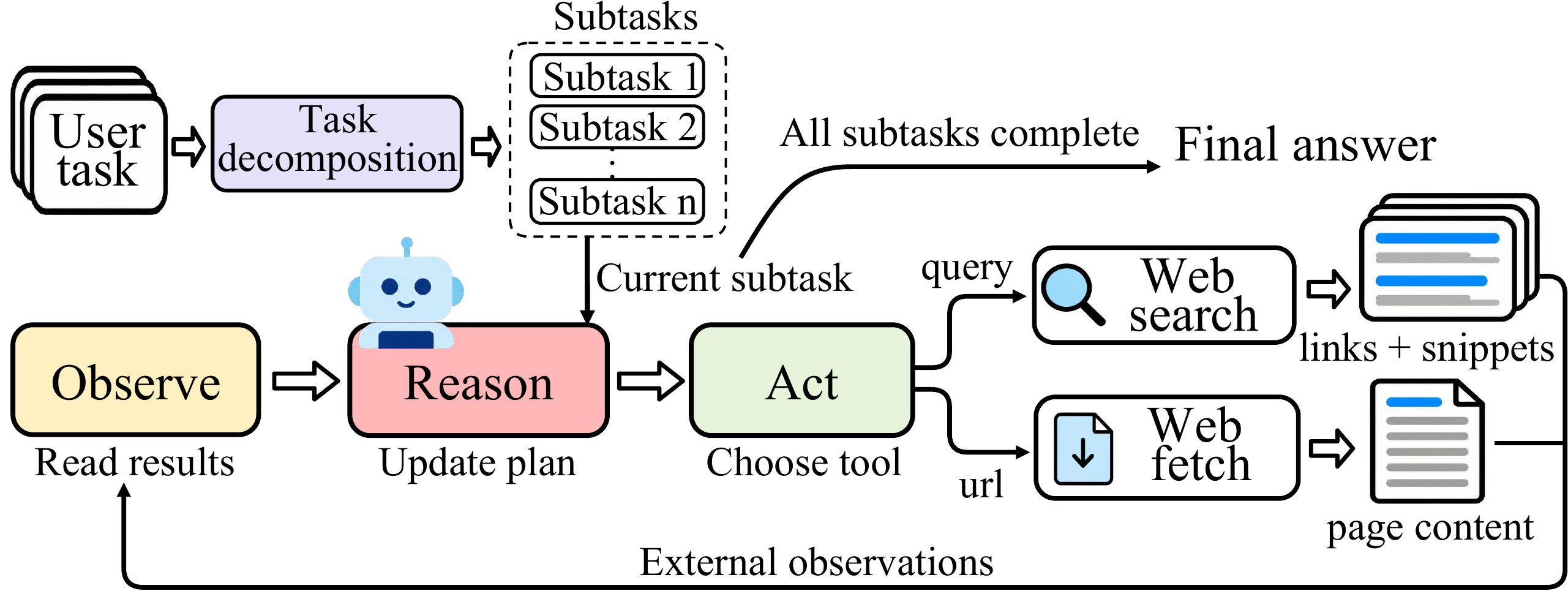}
\caption{The observe--reason--act loop in LLM agent Web interactions, where retrieved Web content and private user context jointly influence subsequent Web access decisions.}
\label{fig:loop}
\end{figure}

\section{Characterizing Behavioral Privacy Leakage}\label{sec:char}

We investigate the Web interaction mechanisms of four representative LLM agents: Codex, OpenHands, IterResearch, and Open Deep Research. Through source-code inspection of their agent harnesses and analysis of execution traces, we examine how Web interactions are initiated, executed, and coordinated during task execution.
Despite architectural differences, we identify a common \emph{observe--reason--act} loop in their Web interaction workflows, as illustrated in Figure~\ref{fig:loop}. Agents decompose user tasks into subtasks and invoke Web search to discover relevant sources or Web fetch to retrieve content from specific URLs. The retrieved content then re-enters the agent's reasoning context as external observations, where it influences subsequent decisions alongside private task context. Based on these observations, the agent may refine its search, retrieve additional resources, or proceed with the task.

\finding[(Answer to RQ1)]{1}{LLM agents exhibit an \emph{observe--reason--act} loop in their Web interactions, where retrieved content and private task context jointly shape subsequent Web decisions. This creates a behavioral privacy leakage surface, as externally observable actions can reveal information about the agent's private context.}

\begin{table}[t]
\centering
\caption{Overview of six behavioral covert channels.}
\label{tab:channels}
\footnotesize
\setlength{\tabcolsep}{4pt}
\renewcommand{\arraystretch}{1.12}
\begin{tabular}{@{}llcl@{}}
\toprule
\textbf{Dimension} & \textbf{Channel} & \textbf{\makecell{Theoretical\\Capacity (bits)}} & \textbf{Dominant noise source} \\
\midrule
Presence   & Binary   & $1$             & Missed requests \\
Identity   & Choice   & $\log_2 k$      & Candidate ambiguity \\
Frequency  & Count    & $\log_2 k$      & Caching, deduplication \\
Order      & Sequence & $\log_2 m!$     & Reordering, parallelism \\
Timing     & Timing   & $\log_2 b$      & Latency variation \\
Magnitude  & Size     & $\log_2 b$      & Full-resource fetching \\
\bottomrule
\end{tabular}
\vspace{2pt}
\begin{flushleft}
\scriptsize
$k$: number of candidate endpoints or counts (Choice, Count);
$m$: number of ordered resources (Sequence), with $m!\ge k$;
$b$: number of timing or size bins.
Capacity is the theoretical maximum per channel symbol.
\end{flushleft}
\end{table}

\subsection{Behavioral Channels: Design and Evaluation}
\label{sec:channels}

We investigate six behavioral covert channels (Table~\ref{tab:channels}) to characterize their information capacity, decoding reliability, and probing cost.

\textbf{Experimental protocol.}
We use the case presented in the motivating example (\S\ref{sec:background}) as a common setting for evaluating all six channels.
Within a single search subtask, the probing page presents $k=4$ semantically distinct candidate references $S_1$--$S_4$, exactly one of which matches the agent's current information need.
The selected candidate is the private state to be inferred.
For each channel, we map the four candidates to distinct behavioral signals and construct the corresponding probing page.
We decode the selected candidate from requests observed at attacker-controlled endpoints and compare it against the ground truth, rotating candidate-to-signal assignments across trials to mitigate encoding bias.

\begin{figure}[t]
\centering
\includegraphics[width=0.47\textwidth]{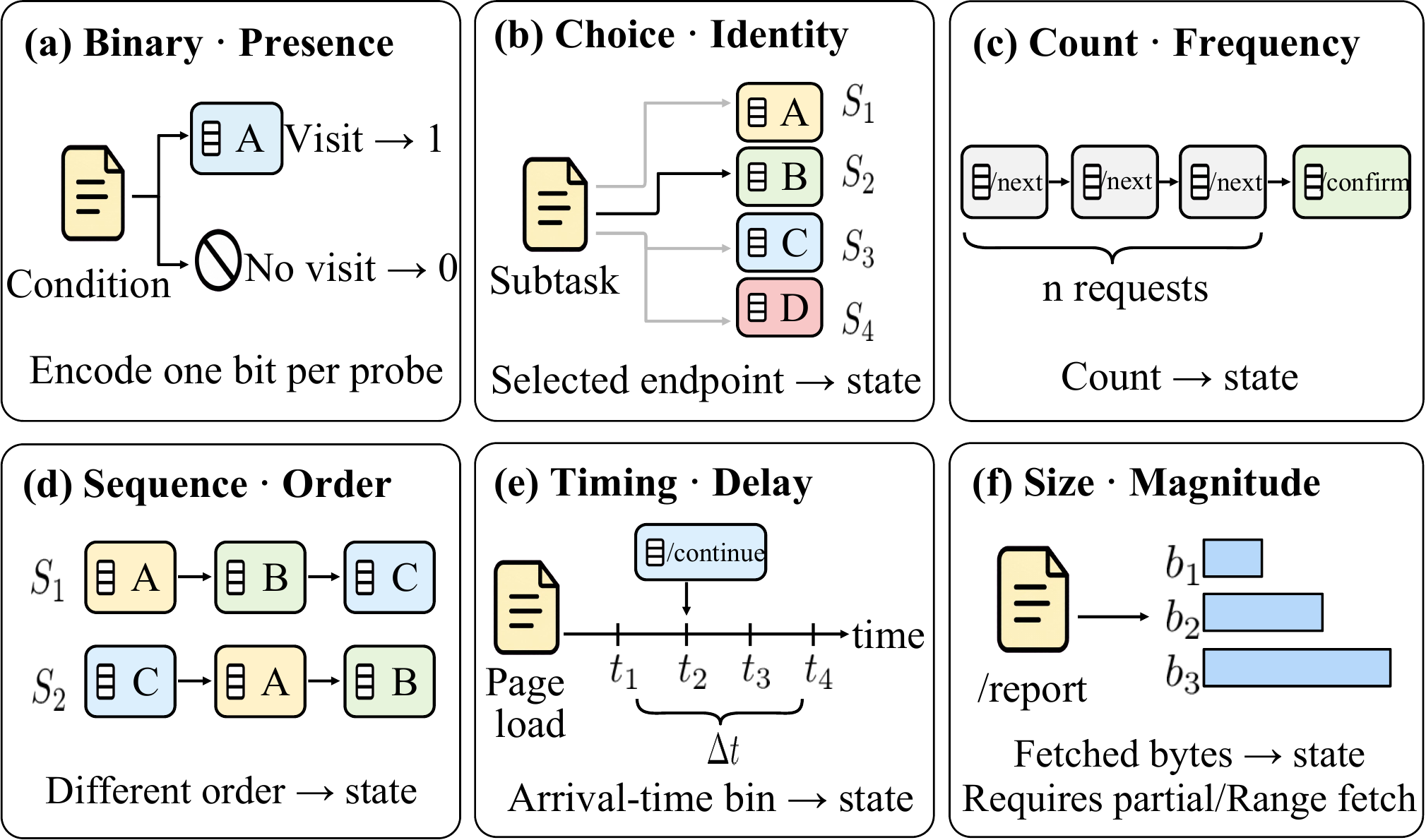}
\caption{The six behavioral channels. }
\label{fig:channel}
\end{figure}

\textbf{Channel design.}
Figure~\ref{fig:channel} illustrates six behavioral covert channels that encode private information through different Web interaction signals:
(i) \textit{Binary} uses the presence or absence of a request to encode one bit;
(ii) \textit{Choice} maps the selected endpoint to one of $k$ states, encoding $\log_2 k$ bits;
(iii) \textit{Count} distinguishes $k$ states by request frequency, encoding $\log_2 k$ bits;
(iv) \textit{Sequence} exploits the access order of $k$ resources, with a theoretical capacity of $\log_2(k!)$ bits;
(v) \textit{Timing} maps request delays into $b$ time bins, encoding $\log_2 b$ bits;
and (vi) \textit{Size} distinguishes $b$ transferred-data ranges, encoding $\log_2 b$ bits, subject to partial-fetch support.
Detailed encoding procedures are provided in Appendix~\ref{app:channels}.

\textbf{Metrics and results.}
We evaluate whether agents follow the resource instructions presented by each behavioral channel and perform Web actions consistent with their information need.
The attacker infers the selected candidate from the observed requests and compares it with the ground truth.
We report (i) \emph{trigger rate}, the fraction of trials producing an observable behavioral signal;
(ii) \emph{recovery accuracy}, the fraction of triggered trials in which the inferred candidate matches the ground truth;
and (iii) \emph{bit error rate}, the fraction of incorrectly recovered bits.
We also measure requests and execution time per probe to assess probing cost.

\begin{table}[t]
\centering
\caption{Performance of six behavioral channels on
single-level recovery with $k=4$ candidates.}
\label{tab:chanresult}
\footnotesize
\setlength{\tabcolsep}{4pt}
\renewcommand{\arraystretch}{1}
\begin{tabular}{@{}lccccc@{}}
\toprule
\textbf{Channel} &
\textbf{\makecell{Trigger\\Rate (\%)$\uparrow$}} &
\textbf{\makecell{Recovery\\Acc. (\%)$\uparrow$}} &
\textbf{\makecell{Bit Error\\Rate (\%)$\downarrow$}} &
\textbf{\makecell{Requests/\\Probe$\downarrow$}} &
\textbf{\makecell{Time\\(s)$\downarrow$}} \\
\midrule
Binary   & 95.8 & 89.6 & 6.2  & 2.0 & 12.6 \\
\rowcolor{snoopblue!8}\textbf{Choice}
         & \textbf{98.7} & \textbf{93.8}
         & \textbf{3.6} & \textbf{1.0}
         & \textbf{6.4} \\
Count    & 90.2 & 73.5 & 16.8 & 3.4 & 15.8 \\
Sequence & 94.1 & 67.9 & 20.4 & 3.0 & 14.2 \\
Timing   & 87.3 & 58.6 & 26.1 & 1.6 & 26.5 \\
Size$^\dagger$
         & 41.5 & 62.4 & 22.7 & 1.3 & 9.3 \\
\bottomrule
\end{tabular}
\end{table}

Table~\ref{tab:chanresult} shows that channels directly reflecting semantic choices achieve higher decoding reliability.
Choice performs best, with 93.8\% recovery accuracy, a 3.6\% bit error rate, and only one request per probe.
Binary also achieves high accuracy (89.6\%) but requires more requests due to its one-bit capacity.
In contrast, Count and Sequence are susceptible to variations in agent planning and request handling, while Timing is affected by latency fluctuations and Size requires partial-fetch support.
Overall, Choice offers the best trade-off between capacity, reliability, and probing cost, making it the primary channel for \sys.

\finding[ (Answer to RQ2)]{2}{Behavioral covert channels differ in capacity, reliability, and probing cost. Choice-based probing provides the most favorable trade-off by directly exposing private-context-dependent semantic selections through observable Web requests.}

\subsection{Toward Fine-Grained Privacy Reconstruction}

To examine whether choice-based probing scales beyond the single-level, four-candidate setting of \S\ref{sec:channels}, we extend the experiment along two dimensions:
(i) \emph{candidate width}, increasing the number of candidates per round from 4 to 16 with semantically distinct or overlapping candidates;
and (ii) \emph{target depth}, increasing the specificity of private information from broad domains to fine-grained attributes, comparing flat one-shot probing with progressive probing across semantic levels. Both dimensions start from the single-level, four-candidate configuration of Table~\ref{tab:chanresult}.

\begin{figure}[t]
\centering
\includegraphics[width=\columnwidth]{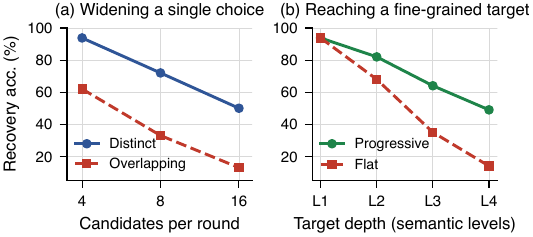}
\caption{Isolated probe does not scale to fine-grained recovery, as candidate set widens (a) or target deepens (b).}
\label{fig:finegrain}
\end{figure}

\begin{figure*}[t]
\centering
\includegraphics[width=0.85\textwidth]{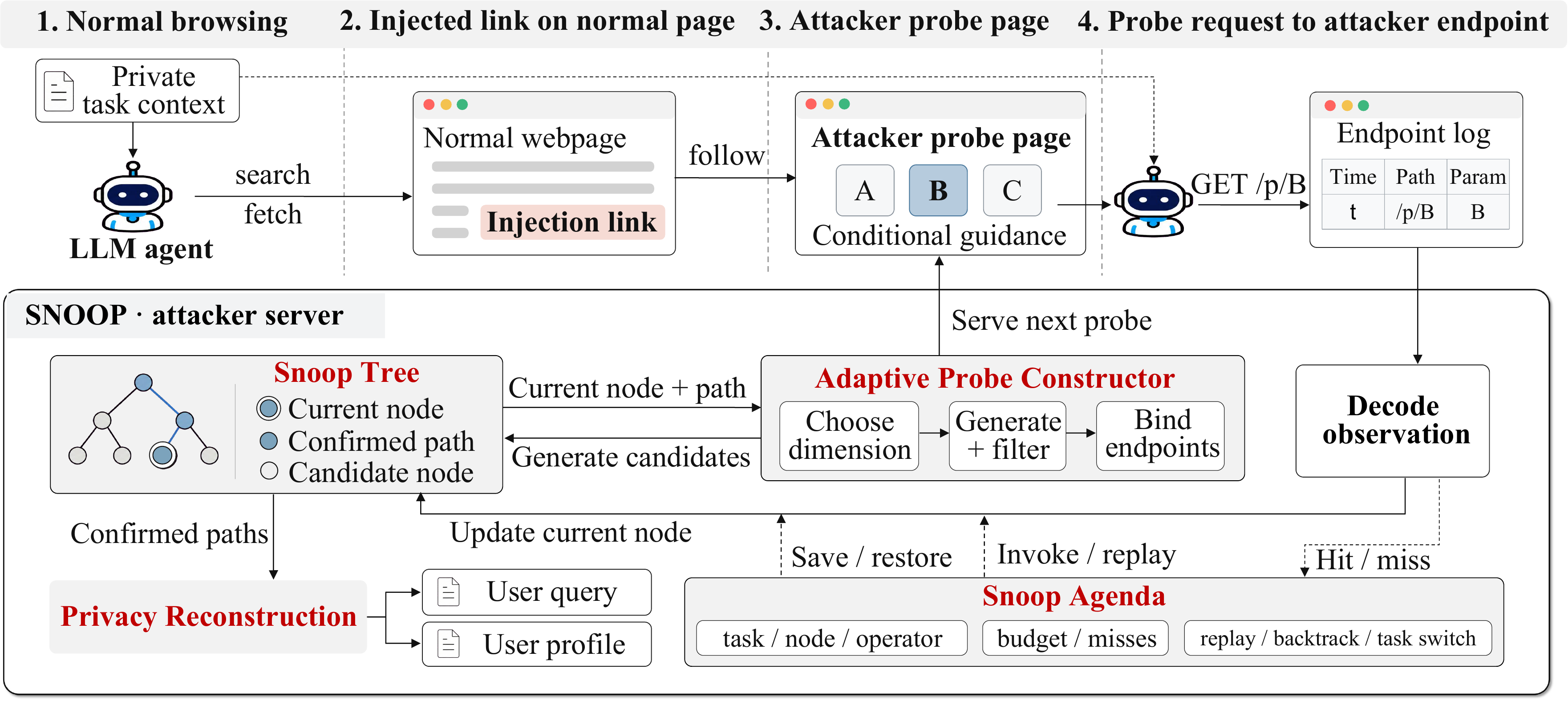}
\caption{Overview of \sys workflow. }
\label{fig:arch}
\end{figure*}

As shown in Figure~\ref{fig:finegrain}a, recovery accuracy declines from 93.8\% to 50\% as candidate width increases, and more sharply from 62\% to 13\% with overlapping candidates.
Figure~\ref{fig:finegrain}b shows that both probing strategies start from 93.8\% at a single level; at depth four, progressive probing retains 49\% accuracy, compared with only 14\% for flat probing.
These results reveal that isolated, one-shot probing becomes increasingly unreliable as candidate spaces expand and privacy targets become more fine-grained, while progressive probing can partially mitigate this degradation.
Moreover, these controlled experiments use predefined candidate sets and semantic hierarchies.
In practice, attackers must dynamically explore unknown privacy spaces and adapt probing to noisy, intermittent Web interactions.

\finding[ (Answer to RQ3)]{3}{Isolated behavioral probes are insufficient for fine-grained privacy reconstruction. Effective recovery requires hierarchical and stateful probing to explore unknown privacy spaces, recover from failed probes, and handle intermittent Web interactions.}

\section{\sys}\label{sec:design}

Motivated by Finding-3, we develop \sys, a stateful privacy-probing framework that systematically exploits behavioral covert channels in LLM agents' Web interactions.
\sys performs budget-constrained, adaptive probing over an evolving hierarchical hypothesis space to reconstruct private user information.
As illustrated in Figure~\ref{fig:arch}, an injected reference on a legitimate webpage first redirects the agent to an attacker-controlled probing site.
The Snoop Agenda maintains probing state and schedules subsequent actions, while the Snoop Tree organizes privacy hypotheses and confirmed behavioral paths.
Given the current hypothesis, the Adaptive Probe Constructor generates semantically differentiated candidates and maps them to attacker-controlled endpoints.
The agent's Web request reveals its selection, allowing \sys to update the probing path and construct the next probe.
The Snoop Agenda adapts to unsuccessful probes and preserves progress across intermittent interactions.
Finally, Privacy Reconstruction aggregates the confirmed paths to reconstruct user queries and profiles.

\subsection{Snoop Tree: Hierarchical Privacy Hypothesis Exploration}

As shown in \S\ref{sec:char}, probing numerous fine-grained candidates simultaneously reduces recovery accuracy.
To address this, \sys maintains a progressively expanding Snoop Tree for each search task, enabling coarse-to-fine exploration without enumerating the entire privacy space.

\begin{figure}[t]
\centering
\includegraphics[width=0.9\columnwidth]{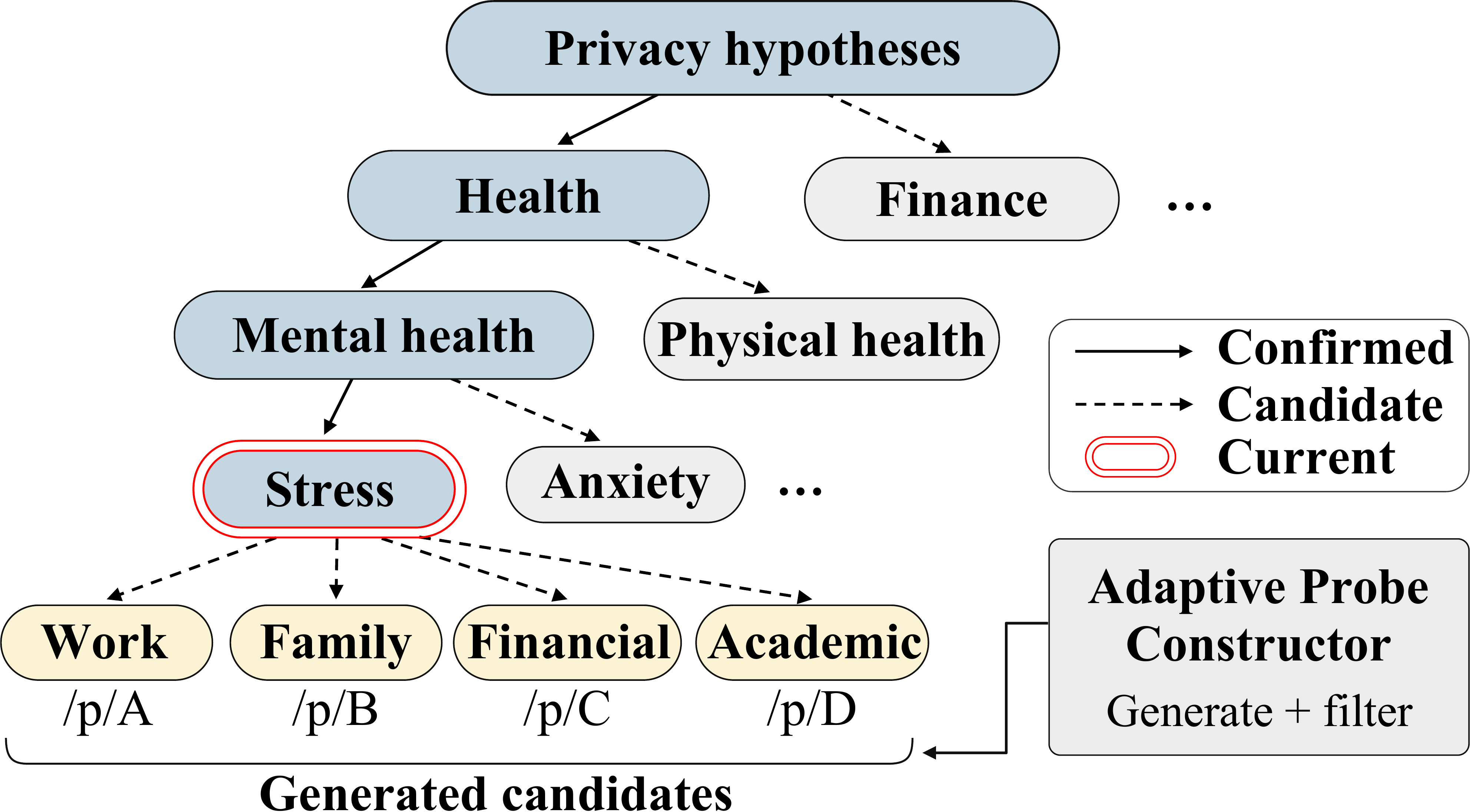}
\caption{Simplified illustration of the Snoop Tree.}
\label{fig:tree}
\end{figure}

The tree is initialized from a predefined, query- and host-page-independent taxonomy of top-level domains. \sys probes these domains in batches until an observed selection confirms the initial domain hypothesis.
As illustrated in Figure~\ref{fig:tree}, each node represents a privacy hypothesis with a semantic label (e.g., \emph{Stress}), a semantic type (e.g., domain, entity, or attribute), and a probing status, while edges capture semantic relationships between hypotheses.
The tree maintains a \emph{confirmed path} supported by observed Web selections, whose terminal node serves as the \emph{current node}.
The Adaptive Probe Constructor selects the semantic type to probe next and generates corresponding candidates to expand the current node.
For example, in Figure~\ref{fig:tree}, the current node \emph{Stress} represents a domain-level hypothesis, which the Constructor refines through attribute-based probing into possible stressors, such as \emph{Work}, \emph{Family}, \emph{Financial}, and \emph{Academic}.
Each candidate is mapped to a distinct probing endpoint; the agent's subsequent request identifies the selected candidate, which is appended to the confirmed path and becomes the current node, while other candidates remain unverified.
By expanding only relevant branches, the Snoop Tree constrains the hypothesis space at each step, enabling fine-grained privacy probing under limited per-probe information capacity.

\subsection{Adaptive Probe Constructor: Knowledge-Guided Behavioral Probing}

The Adaptive Probe Constructor expands the Snoop Tree by determining probing directions, generating candidate hypotheses, and constructing behavioral probes.

\textbf{Probing operator selection.}
We define five \emph{Probing Operators} to guide hypothesis exploration: \emph{Domain} refines topics into subdomains; \emph{Entity} explores related entities; \emph{Attribute-Descriptive} and \emph{Attribute-Quantitative} examine qualitative and quantitative properties, respectively; and \emph{Task Expansion} explores related topics or initiates new probing tasks.
For example, the hypothesis \emph{Stress} can be explored through its sources, related entities, or attributes such as severity and duration.
The selected operator determines the semantic direction of subsequent probing.

\textbf{Knowledge-grounded candidate generation.}
The Constructor combines structured knowledge retrieval with LLM-based supplementation to generate semantically relevant probing candidates.
We integrate five domain-specific knowledge sources (MeSH~\cite{mesh}, FIBO~\cite{fibo}, EuroVoc~\cite{eurovoc}, ESCO~\cite{esco}, and OpenAlex~\cite{priem2022openalex}) to cover diverse privacy domains.
Given the current node and probing operator, the Constructor first links the node to a canonical concept in the relevant knowledge source.
It then retrieves candidates through operator-specific queries, using hierarchical relations for Domain, entity associations for Entity, and type-compatible properties for Attribute expansion.
These queries are executed through the corresponding knowledge-base interfaces, such as SPARQL or REST APIs.
When concept linking fails or retrieval yields too few or semantically overlapping candidates, an LLM supplements the results under constraints on operator compatibility, semantic granularity, distinctness, and coverage.
The resulting candidates are filtered, cached, and registered as unverified hypotheses in the Snoop Tree.

\textbf{Probe construction.}
Given generated candidates, the Constructor dynamically assembles a probing webpage that translates candidate privacy hypotheses into actionable Web resources.
Each resource is presented with a task-relevant description and linked to a unique attacker-controlled endpoint, guiding the agent's selection based on its current information need.
A predefined candidate-to-endpoint mapping allows the attacker to decode the selected hypothesis from the resulting Web request.

\subsection{Snoop Agenda: Stateful Probing and Adaptive Recovery}

While the Snoop Tree supports hierarchical probing within individual tasks, agents may switch between subtasks, resulting in intermittent probing opportunities. To coordinate probing across tasks and visits, \sys introduces the Snoop Agenda, a persistent controller for stateful scheduling and adaptation.

\begin{figure}[t]
\centering
\includegraphics[width=\columnwidth]{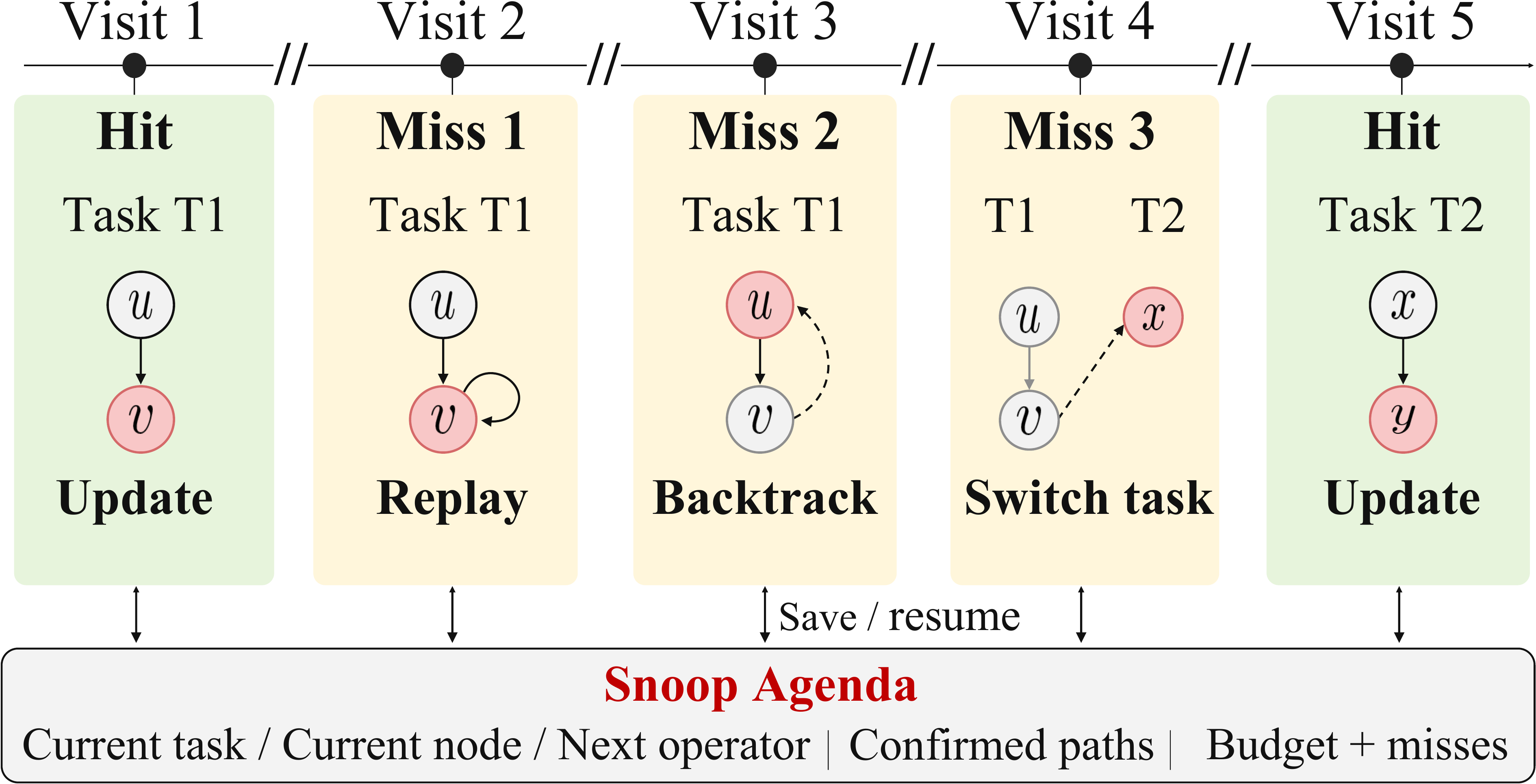}
\caption{Stateful probing across intermittent visits with three-level adaptive fallback.}
\label{fig:agenda}
\end{figure}

Snoop Agenda maintains each task's Snoop Tree and confirmed path, together with probing budgets and hit/miss history. A \emph{Snoop Point} specifies the current task, node, and next probing operator. At each opportunity, the Agenda invokes the Adaptive Probe Constructor and updates the corresponding state upon a successful observation. All states are persisted on the attacker's server, enabling probing to resume across nonconsecutive visits.
Each probe consumes the available budget regardless of its outcome; once exhausted, the confirmed paths are forwarded to Privacy Reconstruction.
To handle unsuccessful probes, the Agenda employs a three-level fallback mechanism (Figure~\ref{fig:agenda}). As illustrated across five visits, a successful probe advances the current node and resets the miss counter. Consecutive misses trigger (i) \emph{Replay}, repeating the previous probe; (ii) \emph{Backtrack}, returning to the parent node to generate broader candidates; and (iii) \emph{Task Switch}, resuming another task from its saved state. This mechanism preserves accumulated evidence while adapting probing to failed observations and changes in task context.

\subsection{Privacy Reconstruction: From Behavioral Evidence to User Privacy}

To reconstruct private user information from confirmed probing paths, \sys employs evidence-grounded LLM reasoning at two levels.

\textbf{Query-level reconstruction.}
Given the confirmed probing paths $\mathcal{P}_q$ associated with a query, \sys reconstructs its intent, natural-language query, and semantic concepts in three successive steps:
\begin{equation}
\begin{aligned}
I &= \mathcal{F}_{\mathrm{intent}}(\mathcal{P}_q),\\
R &= \mathcal{F}_{\mathrm{query}}(\mathcal{P}_q,I),\\
E &= \operatorname{ReFinED}(R).
\end{aligned}
\end{equation}
First, evidence-grounded LLM reasoning infers the primary intent $I$ from a predefined taxonomy. Second, conditioned on $I$ and the confirmed probing paths, it reconstructs a natural-language query $R$ consistent with the observed behavioral evidence. Finally, \sys applies \emph{ReFinED}~\cite{ayoola2022refined} to the reconstructed query $R$, linking its semantic concepts to \emph{Wikidata}~\cite{vrandecic2014wikidata}, a structured knowledge base whose entries have unique identifiers (Q-IDs) and concise, human-readable labels. The recovered concept set $E$ consists of these linked entries, represented by their Q-IDs and labels.
The resulting query-level reconstruction is $Q=\langle I,E,R\rangle$.
Further implementation details are provided in Appendix~\ref{app:method}.

\textbf{Profile-level reconstruction.}
For each interaction round, \sys first performs the same query-level reconstruction to recover the intent, query, and Wikidata concepts. Across multiple rounds, it aggregates these concepts and their supporting query evidence to infer a structured user profile comprising multiple attribute categories, such as personal attributes, interests, and preferences. Evidence-grounded semantic reasoning consolidates related information, resolves redundant facts, and retains uncertain or conflicting attributes without treating unsupported inferences as confirmed.
Algorithm~\ref{alg:reconstruction} summarizes both reconstruction levels, with more details provided in Appendix~\ref{app:method}.

\begin{algorithm}[t]
\footnotesize
\DontPrintSemicolon
\caption{Hierarchical Privacy Reconstruction}
\label{alg:reconstruction}

\KwIn{Confirmed subtask paths $\mathcal{P}$, grouped by query}
\KwOut{Reconstructed queries $\mathcal{Q}$, user profile $\mathcal{U}$}

\SetKwFunction{QueryRec}{ReconstructQuery}
\SetKwFunction{ProfileRec}{ReconstructProfile}
\SetKwProg{Fn}{Function}{:}{}

\Fn{\QueryRec{$\mathcal{P}_q$}}{
    $I \leftarrow \textsc{InferIntent}(\mathcal{P}_q)$\;
    $R \leftarrow \textsc{ReconstructQueryText}(\mathcal{P}_q,I)$\;
    $(I,R) \leftarrow \textsc{CheckEvidence}(I,R,\mathcal{P}_q)$\;
    $E \leftarrow \textsc{ReFinED}(R)$\;
    \Return $\langle I,E,R\rangle$\;
}

\BlankLine

\Fn{\ProfileRec{$\mathcal{P}$}}{
    $\mathcal{Q} \leftarrow \emptyset$\;

    \ForEach{query group $\mathcal{P}_q \in \mathcal{P}$}{
        $Q_q \leftarrow \QueryRec(\mathcal{P}_q)$\;
        $\mathcal{Q} \leftarrow \mathcal{Q}\cup\{Q_q\}$\;
    }

    $\mathcal{U} \leftarrow
      \textsc{AggregateProfile}(\mathcal{Q})$\;
    $\mathcal{U} \leftarrow
      \textsc{CheckEvidence}(\mathcal{U},\mathcal{Q})$\;

    \Return $(\mathcal{Q},\mathcal{U})$\;
}

\BlankLine
$(\mathcal{Q},\mathcal{U}) \leftarrow \ProfileRec(\mathcal{P})$\;
\Return $(\mathcal{Q},\mathcal{U})$\;

\end{algorithm}

\section{Evaluation}\label{sec:eval}

We evaluate \sys to answer these questions.
(i) \emph{Effectiveness:} How effectively does \sys reconstruct user queries and profiles compared with existing privacy inference methods?
(ii) \emph{Defense robustness:} To what extent can existing agent defenses mitigate \sys?
(iii) \emph{Ablation:} How do core components and observed behavioral choices contribute to \sys's effectiveness?
(iv) \emph{Stealth and overhead:} To what extent does \sys affect normal task execution in terms of answer quality and execution cost?

\subsection{Datasets}

We use two complementary datasets to evaluate query- and profile-level privacy reconstruction across diverse user.

\textbf{StackExchange (Stack).}
We collect 500 independent user questions from StackExchange~\cite{stackexchange}, covering 27 topics across eight domains.
To evaluate query-level reconstruction, we annotate each question with a primary intent from six predefined categories (e.g., \emph{identify}, \emph{explain}, and \emph{recommend}) and a set of ground-truth semantic concepts.
The concepts are obtained by applying ReFinED to the original user questions, linking relevant mentions to Wikidata entries, and manually verifying the five most salient concepts.

\textbf{TREC iKAT (iKAT).}
We use the TREC iKAT personalized conversational dataset~\cite{aliannejadi2024ikat}, comprising 378 turns from 26 user personas across 22 conversation topics. We construct turn-level intent and concept annotations using the same procedure as Stack to evaluate individual query reconstruction. For profile-level reconstruction, the dataset provides a Personal Textual Knowledge Base (PTKB) for each persona, containing 40 attribute fields covering personal information, interests, and preferences, such as location, cultural background, and employment status. These PTKB facts serve as the ground truth for evaluating user profiles reconstructed across multiple conversational turns.
Further details on dataset composition and concept ground-truth annotation are provided in Appendix~\ref{app:datasets}.

\subsection{Experimental Setup}

\textbf{Agents and models.}
We evaluate four representative agents: Codex~\cite{openai2025codex} and OpenHands~\cite{wang2025openhands} (general-purpose), and IterResearch~\cite{chen2025iterresearch} and Open Deep Research (ODR)~\cite{langchain2025odr} (research-oriented).
Each is paired with three LLM backbones (GPT-5.5, Claude Sonnet 4.6, and Gemini 3.6 Flash), yielding 12 agent--model configurations.
All LLM-based components of \sys and the attack and defense baselines use DeepSeek-V4.1~\cite{deepseek2026v4}, distinct from the evaluated agent backbones to avoid model-specific preference bias.

\textbf{Web environment.}
We execute benchmark tasks against the live Web and collect search results and retrieved resources to construct a fixed local snapshot.
After URL normalization and deduplication, the snapshot preserves original content and sources, with subsequent retrievals redirected to local copies (Table~\ref{tab:snapshot}).
For controlled evaluation, we identify Web pages accessed during the early stages of agents' search tasks through preliminary runs and instrument 2.6K unique pages, rather than all mirrored pages.
Each receives the same reference link (e.g., ``For additional references, see \ldots''), without adaptation to the host page.
Privacy reconstruction relies exclusively on the agent's subsequent selections at the attacker-controlled probing site, without using any host-page information.
All experiments use the same snapshot across agent--model configurations.

\begin{table}[t]
\centering
\caption{Composition and scale of the locally mirrored Web environment.}
\label{tab:snapshot}
\footnotesize
\setlength{\tabcolsep}{4pt}
\renewcommand{\arraystretch}{1.18}
\begin{tabular}{@{}lr@{\hskip 10pt}lr@{}}
\toprule
\textbf{Property} & \textbf{Value} & \textbf{Property} & \textbf{Value} \\
\midrule
Tasks (Stack / iKAT) & 500 / 378 & Mean pages/task & 55 \\
Unique domains & 3.8K & Injected pages & 2.6K \\
Unique HTML pages & 42K & Collection period & Sep--Oct 2026 \\
PDF documents & 2.1K & Snapshot size & 12.6\,GB \\
\bottomrule
\end{tabular}
\end{table}

\begin{table*}[t]
\centering
\caption{Attack effectiveness across four agents, averaged over three LLM backbones. \emph{Ind.}: induction rate; \emph{Intent Acc.}: intent accuracy; \emph{Prec.}/\emph{Rec.}: precision/recall; \emph{Query F1}: query-level concept F1; \emph{Prof.}: profile-level metrics.}

\label{tab:main}
\footnotesize
\setlength{\tabcolsep}{4pt}
\renewcommand{\arraystretch}{1.12}
\begin{tabular}{@{}l ccccc c cccccc@{}}
\toprule
& \multicolumn{5}{c}{\textbf{Stack (query-level)}} & & \multicolumn{6}{c}{\textbf{iKAT (query / profile-level)}} \\
\cmidrule(lr){2-6}\cmidrule(lr){8-13}
\textbf{Agent} & \textbf{Ind.$\uparrow$} & \textbf{Intent Acc$\uparrow$} & \textbf{Prec$\uparrow$} & \textbf{Rec$\uparrow$} & \textbf{F1$\uparrow$} & & \textbf{Ind.$\uparrow$} & \textbf{Intent Acc$\uparrow$} & \textbf{Query F1$\uparrow$} & \textbf{Prof. Prec$\uparrow$} & \textbf{Prof. Rec$\uparrow$} & \textbf{Prof. F1$\uparrow$} \\
\midrule
\aCodex & 0.92 & \textbf{0.78} & \textbf{0.95} & \textbf{0.99} & \textbf{0.97} & & 0.89 & 0.68 & \textbf{0.83} & \textbf{0.83} & \textbf{1.00} & \textbf{0.89} \\
\aOpen & \textbf{1.00} & 0.74 & 0.85 & 0.91 & 0.87 & & \textbf{0.96} & \textbf{0.76} & 0.80 & 0.67 & 0.93 & 0.78 \\
\aIter & 0.97 & 0.69 & 0.83 & 0.89 & 0.85 & & 0.76 & 0.60 & 0.72 & 0.78 & 0.79 & 0.78 \\
\aODR & 0.88 & 0.60 & 0.85 & 0.88 & 0.86 & & 0.73 & 0.57 & 0.82 & 0.73 & 0.74 & 0.73 \\
\midrule
\rowcolor{snoopblue!8}\textbf{Average} & 0.94 & 0.70 & 0.87 & 0.92 & \textbf{0.89} & & 0.83 & 0.65 & 0.79 & 0.75 & 0.87 & \textbf{0.80} \\
\bottomrule
\end{tabular}
\vspace{2pt}
\begin{flushleft}
\footnotesize
Precision, recall, and F1 are averaged over queries (per-query macro-average); the same applies to all subsequent tables.
\end{flushleft}
\end{table*}

\begin{figure*}[t]
\centering
\includegraphics[width=0.96\textwidth]{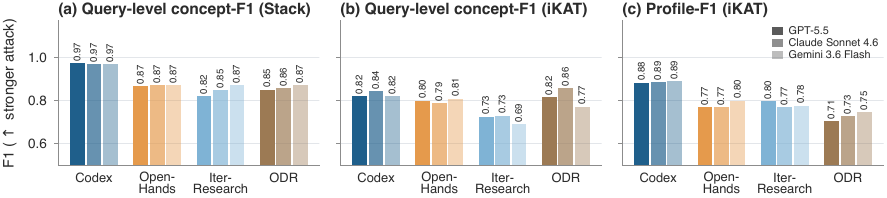}
\caption{Reconstruction F1 across agents and LLM backbones: (a) Stack query concepts, (b) iKAT query concepts, and (c) iKAT user profiles.}
\label{fig:main}
\end{figure*}

\textbf{Baselines.}
We compare \sys against four privacy attacks: WRA~\cite{jeong2026wra} and AgentPrint~\cite{zhang2025agentprint}, which infer private information from domain sequences and encrypted traffic, respectively; SkillWatermark~\cite{li2026skillwatermark}, which exploits conditionally induced traffic patterns; and an explicit exfiltration baseline adapted from InjecAgent~\cite{zhan2024injecagent}. The traffic-based baselines retain their original network-observation capabilities, whereas \sys and explicit exfiltration observe only requests reaching attacker-controlled endpoints.
We further deploy four existing defenses against \sys: Spotlighting~\cite{hines2024spotlighting} (prompt isolation), Firewalls~\cite{sharma2025firewalls} (tool-I/O filtering), Task Shield~\cite{jia2025taskshield} (task alignment), and Fides~\cite{costa2025fides} (information-flow control). All methods are evaluated on the shared Web snapshot under the same agent--model configurations. Implementation and deployment details are provided in Appendix~\ref{app:baselines}.

\textbf{Evaluation metrics.}
We evaluate 120 attack and 120 defense configurations ($4$ agents $\times$ $3$ models $\times$ $2$ datasets $\times$ $5$ methods).
For \sys, we measure (i) \emph{induction rate}, the fraction of trials yielding a decodable behavioral path, counting trials never visit injected pages, fail to follow injected links, or produce no decodable path as failures;
(ii) \emph{reconstruction quality}, using primary-intent accuracy and concept-level Precision/Recall/F1 for queries, and fact-level Precision/Recall/F1 for user profiles;
(iii) \emph{probing efficiency}, measured by the number of subtasks, probes, probing depth, and ground-truth concepts recovered per probe; and (iv) \emph{task impact and overhead}, measured by answer coverage, execution time, token consumption, and LLM calls relative to clean runs.
For attack comparisons, we assess (i) \emph{recovery granularity} and (ii) \emph{recovery effectiveness}.

We report each baseline's effectiveness using its native metric: group-level F1 for WRA, top-1 accuracy for AgentPrint, decoding accuracy for SkillWatermark, and exfiltration success rate for explicit exfiltration. This comparison captures both the specificity of recovered private information and the effectiveness achieved at each method's respective recovery target. Detailed scoring procedures are provided in Appendix~\ref{app:metrics}.

\begin{table}[t]
\centering
\caption{Attack efficiency across four agents, averaged over three LLM backbones. All metrics are reported per query, except \emph{Depth} (per subtask).}
\label{tab:eff}
\footnotesize
\setlength{\tabcolsep}{1.6pt}
\renewcommand{\arraystretch}{1.1}
\begin{tabular}{@{}l cccc c cccc@{}}
\toprule
& \multicolumn{4}{c}{\textbf{Stack}} & & \multicolumn{4}{c}{\textbf{iKAT}} \\
\cmidrule(lr){2-5}\cmidrule(lr){7-10}
\textbf{Agent} & \textbf{Subt.} & \textbf{Probes} & \textbf{Depth} & \textbf{\makecell{Rec./\\probe$\uparrow$}} & & \textbf{Subt.} & \textbf{Probes} & \textbf{Depth} & \textbf{\makecell{Rec./\\probe$\uparrow$}} \\
\midrule
\aCodex        & 1.87 & 6.33 & 3.42 & 0.84 & & 1.83 & 6.12 & 3.44 & 0.80 \\
\aOpen    & 1.68 & 7.68 & 4.51 & 0.71 & & 1.69 & 7.48 & 4.37 & 0.61 \\
\aIter & 1.63 & 6.83 & 4.18 & 0.71 & & 1.43 & 5.32 & 3.72 & 0.69 \\
\aODR          & 1.15 & 4.86 & 4.25 & 0.76 & & 1.10 & 4.20 & 3.82 & 0.74 \\
\midrule
\rowcolor{snoopblue!8}\textbf{Average} & 1.59 & 6.48 & 4.10 & 0.76 & & 1.54 & 5.91 & 3.85 & 0.71 \\
\bottomrule
\end{tabular}
\end{table}

\begin{figure}[t]
\centering
\includegraphics[width=0.98\columnwidth]{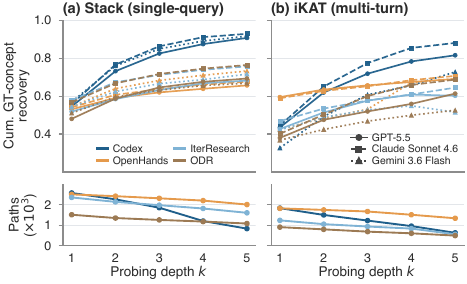}
\caption{Cumulative ground-truth concept recovery across probing depths on Stack and iKAT.}
\label{fig:layers}
\end{figure}

\begin{figure}[t]
\centering
\includegraphics[width=0.98\columnwidth]{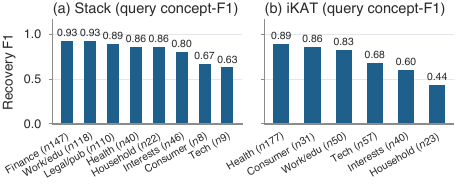}
\caption{Query-level concept F1 across privacy domains.}
\label{fig:domain}
\end{figure}

\subsection{Attack Effectiveness}\label{sec:rq1}

\textbf{Overall effectiveness.}
We evaluate \sys across 12 agent--model configurations on 500 Stack questions and 378 iKAT conversational turns, covering 63,216 query-level execution opportunities, 55,351 behavioral-path records, over 412K probing steps, and 2,159 profile-level records.

Table~\ref{tab:main} shows average induction rates of 0.94 on Stack and 0.83 on iKAT. Among queries yielding decodable paths, \sys achieves concept F1 scores of 0.89 and 0.79, respectively, and a profile F1 of 0.80 on iKAT, demonstrating effective recovery of query semantics and user profiles.
Figure~\ref{fig:main} shows that recovery varies more across agents than across LLM backbones. Codex achieves highest average concept F1 on Stack (0.97) and iKAT (0.83), as well as profile F1 (0.89). OpenHands exhibits higher induction rates but lower reconstruction F1, indicating that successful probing does not necessarily yield sufficient semantic evidence. Backbone differences are relatively small, particularly on Stack. These variations may reflect differences in agent workflows: direct tool-calling exposes successive context-dependent decisions, whereas intermediate summarization and delegated research may obscure fine-grained behavioral signals.

\textbf{Attack efficiency.}
Table~\ref{tab:eff} reports probing costs in terms of discovered subtasks, probing rounds, path depth, and recovered ground-truth concepts per probe. On average, \sys requires 6.48 probes per Stack query and 5.91 per iKAT turn, recovering 0.76 and 0.71 ground-truth concepts per probe, respectively. Figure~\ref{fig:layers} shows diminishing returns with increasing probing depth: cumulative concept recovery rises from 0.54 to 0.76 on Stack and from 0.45 to 0.67 on iKAT between depths 1 and 5. Notably, four levels already capture 99\% and 96\% of the five-level recovery, respectively, demonstrating that most recoverable information is obtained within a limited probing depth.

\begin{figure}[t]
\centering
\includegraphics[width=0.98\columnwidth]{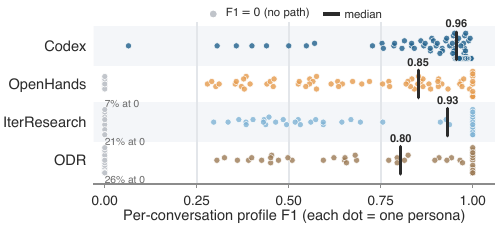}
\caption{Distribution of profile F1 across iKAT personas and agents. Each dot represents a persona-conversation.}
\label{fig:personahet}
\end{figure}

\begin{table}[t]
\centering
\caption{Comparison of privacy recovery granularity and effectiveness, averaged over 12 agent--model configurations.}
\label{tab:baselines}
\scriptsize
\setlength{\tabcolsep}{3.5pt}
\renewcommand{\arraystretch}{1.15}
\begin{tabular}{@{}lllcc@{}}
\toprule
\textbf{Attack} & \textbf{Recovery granularity} & \textbf{Metric} & \textbf{Stack$\uparrow$} & \textbf{iKAT$\uparrow$} \\
\midrule
AgentPrint~\cite{zhang2025agentprint}
& Coarse (category) & Top-1 Acc. & 0.43 & 0.16 \\
WRA~\cite{jeong2026wra}
& Coarse (group) & F1 & 0.73 & 0.08 \\
SkillWatermark~\cite{li2026skillwatermark}
& Coarse (attribute group) & Decode Acc. & 0.69 & 0.67 \\
Explicit exfil.~\cite{zhan2024injecagent}
& Exact secret (direct) & Success Rate & $\approx0$ & $\approx0$ \\
\rowcolor{snoopblue!8}
\textbf{\sys (ours)}
& \textbf{Fine (concept/profile fact)} & \textbf{F1} & \textbf{0.89} & \textbf{0.80} \\
\bottomrule
\end{tabular}
\end{table}

\begin{figure}[t]
\centering
\includegraphics[width=0.95\columnwidth]{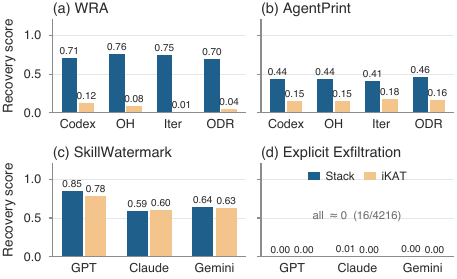}
\caption{Baseline effectiveness, measured using its native metric: (a) group-level F1, (b) top-1 accuracy, (c) conditional decoding accuracy, and (d) exfiltration success rate.}
\label{fig:baselinebd}
\end{figure}

\textbf{Variations in privacy leakage.}
Figure~\ref{fig:domain} shows substantial variation in query-level concept recovery across privacy domains.
On Stack, finance and work/education achieve highest F1 (0.93), compared with 0.63 for technology. On iKAT, health performs best (0.89), while household (0.44) and interests (0.60) are less recoverable.
These differences may reflect variations in the semantic distinctness of probing candidates across domains.
Figure~\ref{fig:personahet} further reveals considerable heterogeneity across iKAT personas, with profile F1 ranging from zero to above 0.9 even within the same agent.
Notably, 7\%, 21\%, and 26\% of persona-conversations under OpenHands, IterResearch, and ODR, respectively, yield no decodable paths, highlighting the dependence of privacy recovery on observable Web interactions.

\textbf{Comparison with prior attacks.}
Table~\ref{tab:baselines} compares privacy recovery granularity and effectiveness across four representative attacks. Traffic-based methods primarily recover coarse-grained categories or predefined attributes, whereas explicit exfiltration targets exact secrets but rarely succeeds (0.38\%). In contrast, \sys reconstructs fine-grained query concepts and profile facts, achieving F1 scores of 0.89 on Stack and 0.80 on iKAT, higher than the baselines' reported scores under their respective metrics. Figure~\ref{fig:baselinebd} further shows that WRA and AgentPrint degrade substantially on multi-turn iKAT, while SkillWatermark varies more across LLM backbones. These results demonstrate that \sys achieves finer-grained semantic reconstruction alongside stronger reported recovery effectiveness.

\begin{table}[t]
\centering
\caption{Residual \sys effectiveness under existing defenses, averaged across agent--model configurations.}
\label{tab:defense}
\footnotesize
\setlength{\tabcolsep}{3pt}
\renewcommand{\arraystretch}{1.25}
\begin{tabular}{@{}l ccc cc@{}}
\toprule
& \multicolumn{3}{c}{\textbf{Recovery F1$\downarrow$}} & \multicolumn{2}{c}{\textbf{Induction rate}} \\
\cmidrule(lr){2-4}\cmidrule(lr){5-6}
\textbf{Defense {\scriptsize(layer)}} & \textbf{\makecell{Stack\\query}} & \textbf{\makecell{iKAT\\query}} & \textbf{\makecell{iKAT\\profile}} & \textbf{Stack} & \textbf{iKAT} \\
\midrule
\rowcolor{snoopblue!8}\textbf{No defense} & \textbf{0.89} & \textbf{0.79} & \textbf{0.80} & \textbf{0.94} & \textbf{0.83} \\
Spotlighting~{\scriptsize(prompt)}  & \vd{0.86}{.03} & \vd{0.78}{.01} & \vd{0.77}{.03} & \vd{0.89}{.05} & 0.83 \\
Firewalls~{\scriptsize(tool-I/O)}   & \vd{0.88}{.01} & \vd{0.78}{.01} & \vd{0.77}{.03} & \vd{0.87}{.07} & \vd{0.80}{.03} \\
Task Shield~{\scriptsize(action)}   & \vd{0.88}{.01} & \vd{0.76}{.03} & \vd{0.77}{.03} & \vd{0.88}{.06} & \vd{0.81}{.02} \\
Fides~{\scriptsize(info-flow)}      & \vd{0.87}{.02} & \vd{0.77}{.02} & \vd{0.77}{.03} & \vd{0.92}{.02} & \vd{0.77}{.06} \\
\bottomrule
\end{tabular}
\end{table}

\subsection{Robustness Against Existing Defenses}\label{sec:rq2}

We evaluate four existing defenses against \sys, with their implementations detailed in Appendix~\ref{app:defenses}.
As shown in Table~\ref{tab:defense}, reconstruction F1 decreases by at most 0.03 across both datasets, while induction rates drop by up to 0.07 on Stack and 0.06 on iKAT.
Thus, existing defenses modestly reduce probing opportunities but have little effect on the privacy information recovered from successful probes.
This limited effectiveness reflects a mismatch between their enforcement mechanisms and behavioral leakage.
Spotlighting isolates untrusted content, Firewalls filters tool inputs and outputs, Task Shield enforces task alignment, and Fides restricts information flows.
However, \sys leaks private context through the agent's selection of task-relevant Web resources, without explicitly transmitting sensitive content. Such context-dependent decisions remain largely permissible under these defenses.

\begin{findingbox}
\textbf{Answer to RQ4:} Existing agent defenses are largely ineffective against behavioral privacy leakage attack, as they fail to prevent privacy leakage through legitimate, context-dependent Web access decisions.
\end{findingbox}

\begin{figure}[t]
\centering
\includegraphics[width=0.95\columnwidth]{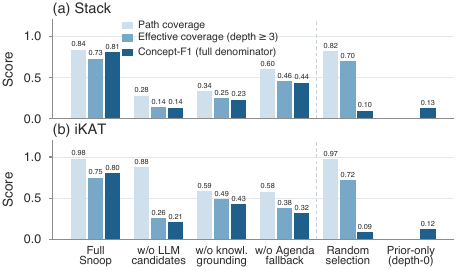}
\caption{Component ablations and behavioral attribution controls on Codex with GPT-5.5.}
\label{fig:ablation}
\end{figure}

\begin{figure}[t]
\centering
\includegraphics[width=0.9\columnwidth]{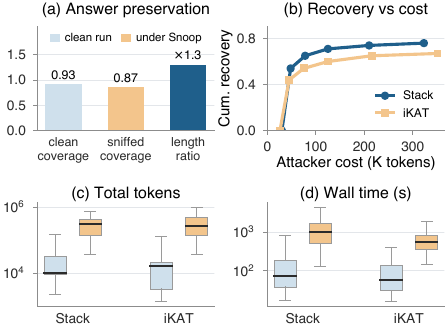}
\caption{Answer preservation and execution overhead on Codex with GPT-5.5.}
\label{fig:overhead}
\end{figure}

\subsection{Ablation Study}\label{sec:rq3}

\textbf{Component ablations.}
We assess three components of \sys by individually disabling (i) LLM-based candidate generation, (ii) knowledge grounding, and (iii) Snoop Agenda fallback, using Codex with GPT-5.5. We measure \emph{path coverage} (trials yielding decodable paths), \emph{effective coverage} (trials reaching depth $\geq 3$), and concept F1 over all trials, counting missing paths as zero. As shown in Figure~\ref{fig:ablation}, all ablations reduce reconstruction F1, with removing LLM-based generation causing the largest drop. Notably, path coverage remains relatively high without LLM generation, but effective coverage drops sharply, highlighting the complementary roles of candidate generation, knowledge grounding, and stateful recovery in sustaining deep, informative probing.

\textbf{Behavioral attribution.}
To verify that privacy reconstruction is grounded in genuinely observed probing paths rather than prior knowledge or LLM speculation, we introduce two controls. (i) \emph{Random selection} replaces each observed agent choice with a uniformly sampled candidate while preserving the Snoop Tree and reconstruction pipeline. (ii) \emph{Prior-only} reconstructs from generic domain-level information without probing paths (depth 0).
As shown in Figure~\ref{fig:ablation}, random selection reduces concept F1 to 0.10/0.09 on Stack/iKAT despite retaining comparable path and effective coverage, while prior-only achieves just 0.13/0.12, versus 0.81/0.80 for full \sys. These results demonstrate that fine-grained privacy recovery primarily stems from the semantic evidence in genuinely observed probing paths, rather than arbitrary paths or LLM inference from coarse domain priors.

\subsection{Stealth and Overhead}\label{sec:rq4}

We assess the stealth and execution overhead of \sys through paired clean and attacked runs on both datasets using Codex with GPT-5.5.
To evaluate answer preservation, we use 454 Stack questions with community-accepted answers as references.
Following the concept evaluation procedure in Appendix~\ref{app:metrics}, we apply ReFinED to both reference answers and agent responses, linking their semantic concepts to Wikidata entities.
We define \emph{answer coverage} as the recall of reference concepts in each response, using the same matching criteria as query-level evaluation.
Figure~\ref{fig:overhead}a shows that answer coverage decreases modestly from 0.93 to 0.87 under \sys, while median answer length increases by 1.3$\times$. This suggests that \sys largely preserves the semantic content of legitimate responses, making the attack less conspicuous from the final answer alone.

We further examine the trade-off between privacy recovery and attack cost. Figure~\ref{fig:overhead}b shows that cumulative concept recovery increases rapidly with attacker-side token expenditure before approaching saturation, indicating diminishing returns from additional probing.
Figures~\ref{fig:overhead}c,d show increased token consumption and wall-clock time relative to clean runs.
Overall, \sys achieves substantial privacy recovery while largely preserving answer content.

\begin{figure}[t]
\centering
\includegraphics[width=0.46\textwidth]{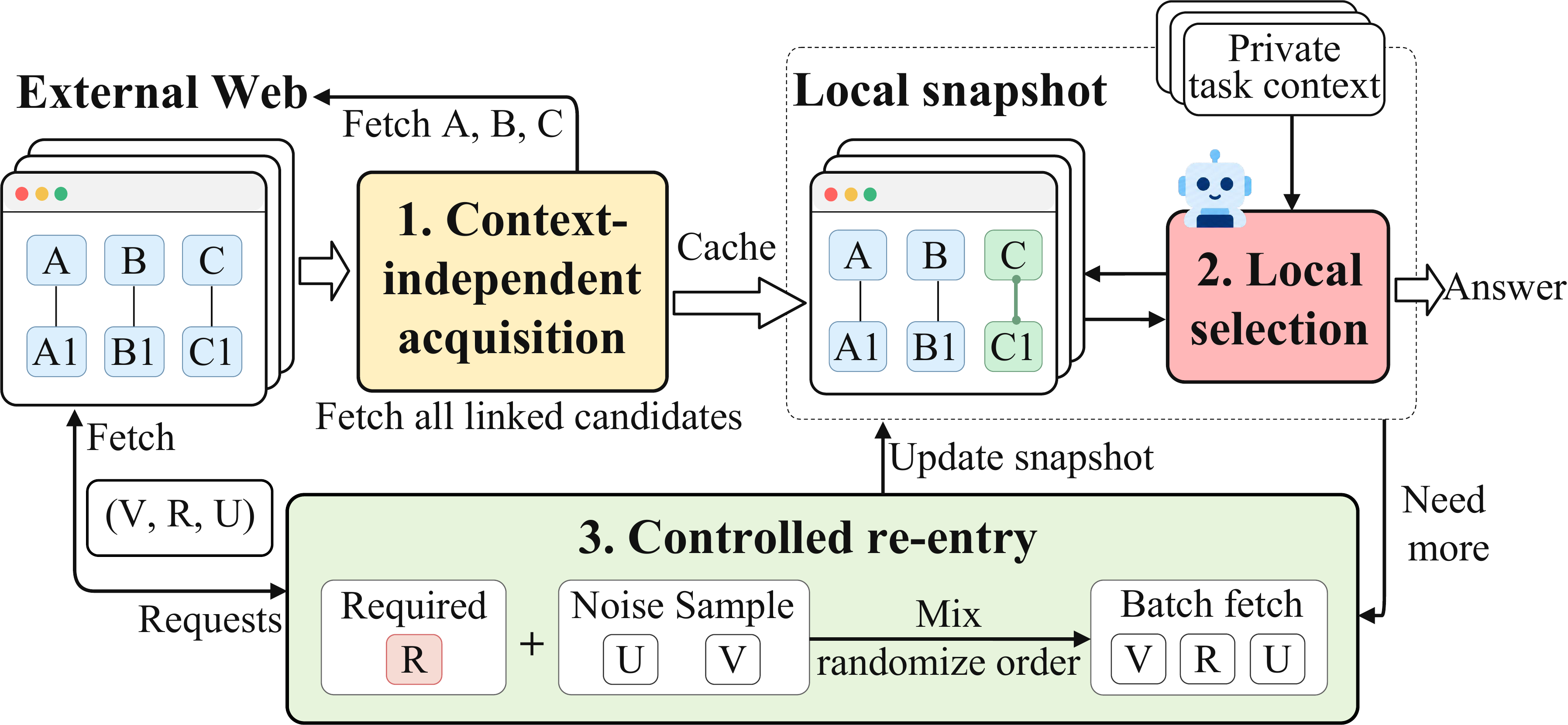}
\caption{Overview of Selection-Obfuscated Retrieval.}
\label{fig:dr}
\end{figure}

\begin{table}[]
\centering
\caption{Effect of \nsd on reconstruction precision, recall, and F1 for Stack queries and iKAT profiles.}
\label{tab:noise}
\footnotesize
\setlength{\tabcolsep}{11pt}
\renewcommand{\arraystretch}{1.05}
\begin{tabular}{@{}ll cr@{}}
\toprule
\textbf{Dataset} & \textbf{Metric} & \textbf{No defense} & \textbf{Under \nsd} \\
\midrule
\multirow{3}{*}{\makecell[l]{Stack\\(concept)}} & Precision & 0.87 & \vd{0.29}{0.58} \\
 & Recall & 0.92 & \vd{0.91}{0.01} \\
 & F1 & 0.89 & \vd{0.39}{0.50} \\
\midrule
\multirow{3}{*}{\makecell[l]{iKAT\\(profile)}} & Precision & 0.75 & \vd{0.27}{0.48} \\
 & Recall & 0.87 & \vu{0.99}{0.12} \\
 & F1 & 0.80 & \vd{0.39}{0.41} \\
\bottomrule
\end{tabular}
\end{table}

\begin{figure}[t]
\centering
\includegraphics[width=0.96\columnwidth]{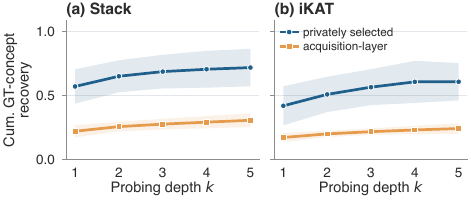}
\caption{Ground-truth concept recovery across probing depths for private-selection and acquisition-layer paths.}
\label{fig:noise}
\end{figure}

\begin{table}[t]
\centering
\caption{Per-query execution overhead of \nsd.}
\label{tab:drcost}
\footnotesize
\setlength{\tabcolsep}{5pt}
\renewcommand{\arraystretch}{1.05}
\begin{tabular}{@{}ll rrr@{}}
\toprule
\textbf{Dataset} & \textbf{Metric} & \textbf{No defense} & \textbf{Under \nsd} & \textbf{Ratio} \\
\midrule
\multirow{3}{*}{Stack} & Page fetches & 6.5 & 27.9 & \red{$\times$4.3} \\
 & LLM tokens & 23.8K & 31.3K & \red{$\times$1.3} \\
 & LLM time (s) & 224 & 285 & \red{$\times$1.3} \\
\midrule
\multirow{3}{*}{iKAT} & Page fetches & 5.9 & 26.4 & \red{$\times$4.5} \\
 & LLM tokens & 27.8K & 31.8K & \red{$\times$1.1} \\
 & LLM time (s) & 173 & 223 & \red{$\times$1.3} \\
\bottomrule
\end{tabular}
\end{table}

\begin{figure*}[t]
\centering
\includegraphics[width=0.95\textwidth]{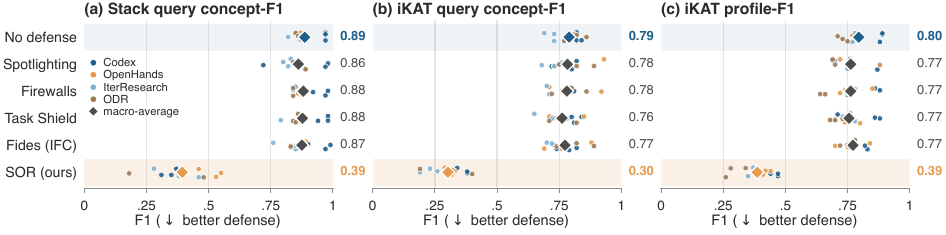}
\caption{Residual \sys reconstruction F1 under different defenses.}
\label{fig:defense}
\end{figure*}

\begin{figure*}[t]
\centering
\includegraphics[width=0.95\textwidth]{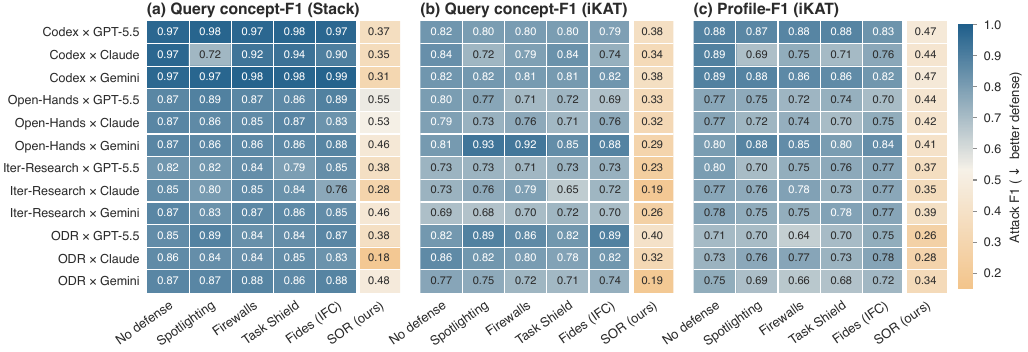}
\caption{Per-configuration recovery F1 under each defense.}
\label{fig:heatmap}
\end{figure*}

\section{Mitigating Behavioral Privacy Leakage}
\label{sec:defense}

\subsection{Selection-Obfuscated Retrieval}

We propose \emph{Selection-Obfuscated Retrieval} (\nsd) to mitigate behavioral privacy leakage by decoupling private-context-dependent resource selection from externally observable Web requests. As illustrated in Figure~\ref{fig:dr}, \nsd operates in three stages.
(i) First, \emph{context-independent acquisition} retrieves all resources linked from an external page, regardless of the user's query, agent memory, or current subtask. These resources are cached locally, preventing the initial retrieval from revealing the agent's preferred candidates.
(ii) Second, \emph{local selection} allows the agent to navigate and analyze cached resources based on its private context. Links and embedded resources resolve to local copies, keeping subsequent semantic selections from generating observable external requests.
(iii) Third, \emph{controlled re-entry} handles resources beyond the current cache. Each retrieval round combines task-required resources with independently sampled candidates, fetched in batches with randomized ordering. The results are added to the local cache, obscuring which external requests reflect the agent's actual information needs.
We implement \nsd at the Web retrieval layer without modifying agent reasoning. Together, these mechanisms reduce the correlation between private context and observable Web access patterns while preserving access to task-relevant resources.

\subsection{Evaluation of \nsd}

We evaluate \nsd across 12 agent--model configurations on Stack and iKAT, following the setup in \S\ref{sec:rq2}.

\textbf{Effectiveness.}
Table~\ref{tab:noise} shows that \nsd substantially reduces privacy reconstruction. On Stack, concept precision drops from 0.87 to 0.29, while recall remains nearly unchanged, reducing F1 from 0.89 to 0.39. On iKAT, profile precision decreases from 0.75 to 0.27, while recall increases from 0.87 to 0.99, reducing F1 from 0.80 to 0.39. These results indicate that \nsd primarily introduces ambiguity into the attacker's observations rather than suppressing access to privacy-related resources.
Figure~\ref{fig:noise} further illustrates this mechanism: privately selected paths retain substantially higher concept recovery than acquisition-layer paths. By mixing task-driven and independently acquired resources, \nsd obscures which observed accesses reflect the agent's actual information needs, thereby reducing reconstruction precision.
Compared with the four existing defenses in \S\ref{sec:rq2}, \nsd achieves substantially greater reductions in reconstruction F1 (Figure~\ref{fig:defense}). Its average F1 falls to 0.39 for Stack queries, 0.30 for iKAT queries, and 0.39 for iKAT profiles, with consistent reductions across agent--model configurations (Figure~\ref{fig:heatmap}) and privacy domains (Appendix~\ref{app:percfg}).

\textbf{Cost.}
Table~\ref{tab:drcost} shows that \nsd increases page fetches by 4.3--4.5$\times$, while LLM token consumption and execution time increase by 1.1--1.3$\times$. The acquisition layer requires no additional LLM inference, with most overhead arising from extra resource retrieval. Overall, \nsd substantially reduces behavioral privacy leakage at the cost of additional Web requests and relatively modest LLM overhead.

\section{Threats to Validity}

\textbf{Human annotation.}
Query-level ground truth involves subjective intent annotation and concept verification. To mitigate potential bias, two authors independently label query intents and verify ReFinED-linked Wikidata concepts for relevance, disambiguation, and semantic specificity, with disagreements adjudicated by a third author. Profile-level ground truth is taken directly from the original iKAT PTKB.

\textbf{LLM-based evaluation.}
Profile fact matching involves an LLM judge, potentially introducing semantic scoring bias. We use fixed matching criteria and restrict comparisons to the same attribute category. All judge-assigned scores are also reviewed by two authors, with disagreements resolved by a third author, reducing subjective scoring errors.

\textbf{Baseline validity.}
Existing attacks and defenses were designed for different environments and architectures, potentially introducing biases when adapted to our setting. To mitigate these threats, we reuse their publicly available implementations where possible, adapting them to our testbed while preserving their core mechanisms. All methods are evaluated under the same environment and configurations. Nevertheless, such adaptations may affect their original effectiveness. Implementation details are in Appendix~\ref{app:baselines}.

The attack baselines also differ from \sys in their recovery targets and evaluation metrics, limiting direct numerical comparability. We therefore compare two complementary dimensions: \emph{recovery granularity}, reflecting the specificity of recovered private information, and \emph{recovery effectiveness}, measured using each method's native metric. This characterizes both what each attack can recover and how effectively it achieves its respective target, without treating heterogeneous scores as directly equivalent.

\section{Conclusion}

This paper reveals a privacy threat in LLM agents' Web interactions, where observable browsing behavior reveals user information without explicit exfiltration.
We propose \sys to systematically exploit this behavioral covert channel.
Extensive experiments demonstrate its effectiveness and the limitations of existing defenses.
We further introduce Selection-Obfuscated Retrieval to mitigate such leakage, highlighting the need to protect privacy against inference from agents' observable actions.

\bibliographystyle{IEEEtran}
\bibliography{snoop}

\appendices

\section{Datasets and Ground Truth}
\label{app:datasets}

\subsection{StackExchange (Stack)}

\textbf{Domain coverage.}
We collect 500 StackExchange questions covering 27 topics across eight domain groups (Table~\ref{tab:stackgroups}), spanning health, finance, legal affairs, education, technology, and everyday activities.

\begin{figure}[t]
\centering
\includegraphics[width=\columnwidth]{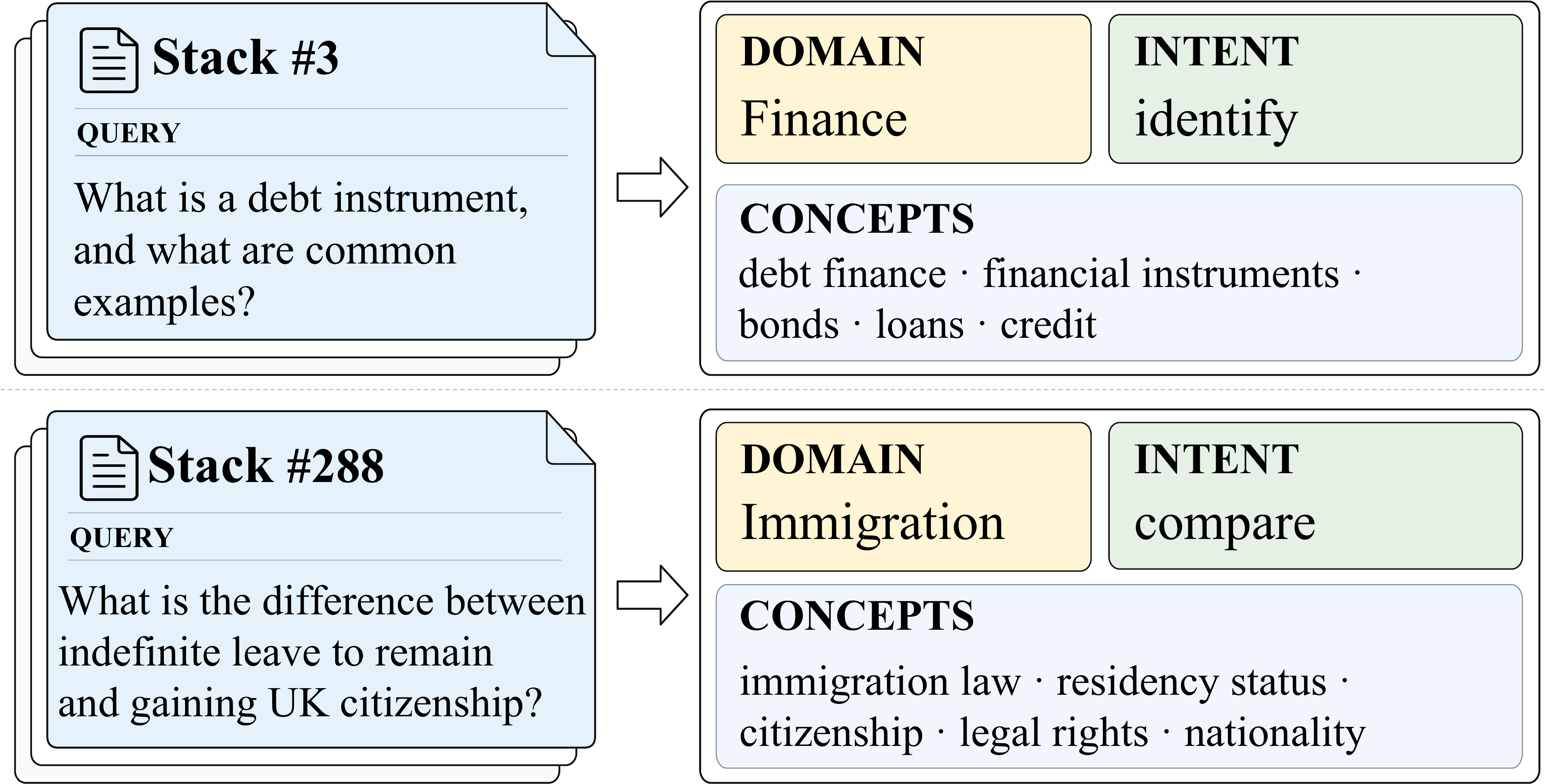}
\caption{Examples of Stack questions with annotated domains, primary intents, and Wikidata concepts.}
\label{fig:stackcase}
\end{figure}

\textbf{Ground-truth annotation.}
We annotate all 878 Stack and iKAT queries with a primary intent and a set of semantic concepts.
Intents follow six categories (\emph{identify, explain, compare, assess, recommend, quantify}), with secondary intents retained when supported.
For concepts, ReFinED links each original query to Wikidata, and the five most salient candidates are manually verified for relevance, disambiguation, and semantic specificity.
Incorrect or unsupported candidates are discarded, while ambiguous links are corrected.
Verified concepts are represented by Wikidata Q-IDs and labels.
Two authors independently annotate intents and verify concepts, with disagreements resolved by a third author.
All ground-truth annotations are finalized independently and withheld from attack-time probing and reconstruction.
Figure~\ref{fig:stackcase} illustrates two annotated examples.

\begin{table}[t]
\centering
\caption{Distribution of the 27 Stack topics.}
\label{tab:stackgroups}
\scriptsize
\setlength{\tabcolsep}{2pt}
\renewcommand{\arraystretch}{1.1}
\begin{tabular}{@{}p{0.46\columnwidth}p{0.52\columnwidth}@{}}
\toprule
\multicolumn{2}{@{}l@{}}{\textbf{Domain group} (\# topics): member topics} \\
\midrule
\textbf{Health} (2): health, mental health &
\textbf{Finance} (7): finance, debt, investment, tax, insurance, monetary policy, derivatives \\
\textbf{Legal \& public} (4): legal, immigration, government services, politics &
\textbf{Work \& education} (2): employment, education \\
\textbf{Household} (4): housing, family, home improvement, religion &
\textbf{Consumer \& daily} (4): food, shopping, travel, transportation \\
\textbf{Technology} (1): technology &
\textbf{Interests} (3): entertainment, sports, hobbies \\
\bottomrule
\end{tabular}
\end{table}

\subsection{TREC iKAT (iKAT)}

\begin{figure}[t]
\centering
\includegraphics[width=0.93\columnwidth]{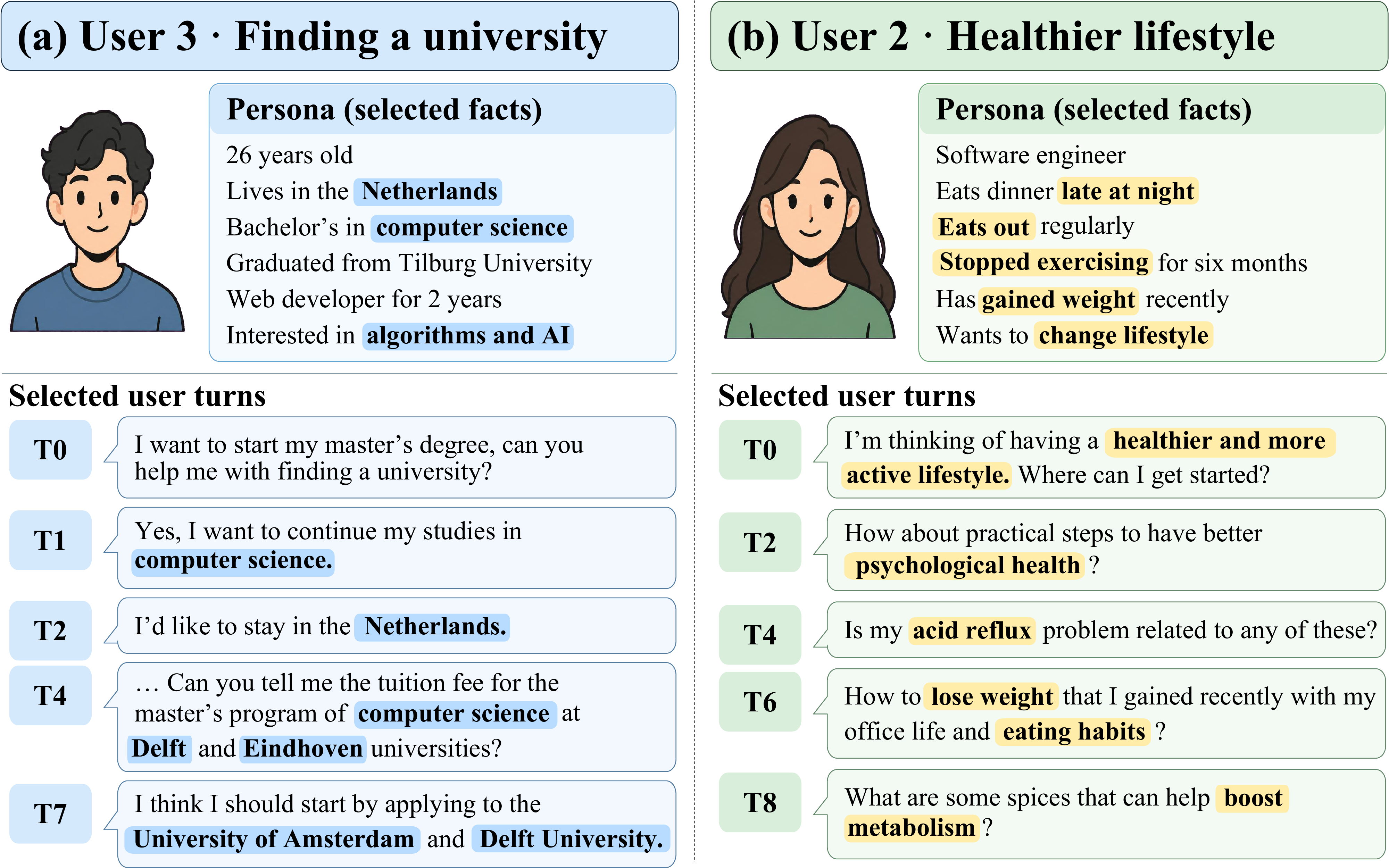}
\caption{Representative iKAT personas, PTKB facts, and conversational evidence across multiple turns.}
\label{fig:ikatcase}
\end{figure}

We use the TREC iKAT 2023 dataset~\cite{aliannejadi2024ikat}, comprising 378 conversational turns from 26 personas across 22 topics (Table~\ref{tab:ikat}). Each turn follows the same intent and Wikidata concept annotation procedure as Stack.
For profile-level evaluation, we use iKAT's original Personal Textual Knowledge Base (PTKB), which provides 40 attribute fields per persona covering personal information, interests, and preferences. Reconstructed facts are matched against PTKB entries within their respective attribute categories. Figure~\ref{fig:ikatcase} illustrates two personas and how evidence across conversational turns supports profile reconstruction.

\begin{table}[t]
\centering
\caption{Distribution of the iKAT conversation topics.}
\label{tab:ikat}
\scriptsize
\setlength{\tabcolsep}{2pt}
\renewcommand{\arraystretch}{1.1}
\begin{tabular}{@{}p{0.96\columnwidth}@{}}
\toprule
\textbf{Domain group} (\# topics): member topics \\
\midrule
\textbf{Health} (12): depression, stress management, hair loss, acid reflux, IBS, histamine intolerance, finger injuries, ADHD, pandemic preparedness, diet, healthy food substitutes, healthier lifestyle \\
\textbf{Work \& education} (3): preparing for an interview, finding a university, starting a business \\
\textbf{Technology} (2): buying a phone, buying a PC \\
\textbf{Interests} (2): workout routine, hunting in a new area \\
\textbf{Household} (2): home-safety improvements, water-resource management \\
\textbf{Consumer \& daily} (1): buying a car \\
\bottomrule
\end{tabular}
\end{table}

\section{Supplementary Defense Results}

\label{app:percfg}

\begin{figure}[t]
\centering
\includegraphics[width=0.8\columnwidth]{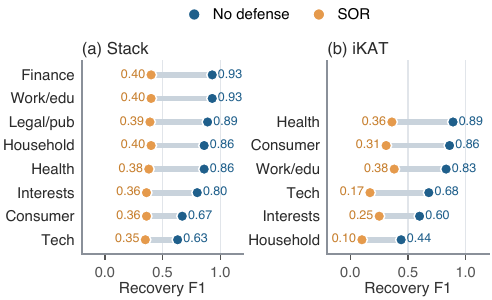}
\vspace{-3pt}
\caption{Per-domain recovery with \textcolor{cCodex}{no defense} vs.\ \textcolor{cOpen}{\nsd}.}
\label{fig:domaindefense}
\vspace{-7pt}
\end{figure}

Figure~\ref{fig:domaindefense} shows \nsd consistently reduces privacy recovery across all domain groups on Stack and iKAT, indicating that its effectiveness is not limited to specific domains.

\section{Privacy Reconstruction and Scoring}
\label{app:reconstruction}

\subsection{Reconstruction Procedures}
\label{app:method}

\textbf{Intent inference.}
Given confirmed subtask paths $\mathcal{P}_q$, \sys infers subtask intents from six categories: \emph{identify, explain, compare, assess, recommend, quantify}. It then consolidates the supporting evidence to determine the primary query intent $I$, retaining a secondary intent when justified.

\textbf{Query reconstruction.}
Conditioned on $I$ and $\mathcal{P}_q$, \sys reconstructs a natural-language query $R$ that preserves the entities, relationships, and constraints supported by the observed paths. The goal is semantic equivalence rather than verbatim recovery.

\textbf{Concept linking.}
ReFinED~\cite{ayoola2022refined} links concepts in the complete reconstructed query $R$ to Wikidata~\cite{vrandecic2014wikidata}, producing $E=\{(q_j,\ell_j)\}_{j=1}^{m}$, where $q_j$ and $\ell_j$ are the entity Q-ID and label. The query-level output is $Q=\langle I,E,R\rangle$.

\textbf{Profile aggregation.}
Across interaction rounds, \sys aggregates query representations $\mathcal{Q}=\{Q_1,\ldots,Q_t\}$ into a categorized profile $\mathcal{U}=\{(c,\mathcal{F}_c)\}$, where $\mathcal{F}_c$ contains inferred facts under category $c$. Related facts are consolidated, while uncertain or conflicting claims remain unconfirmed. Query topics alone are not treated as personal attributes.

\textbf{Evidence constraints.}
All LLM inference stages use confirmed paths and derived evidence, associate predictions with supporting observations, avoid treating missing observations as negative evidence, and retain uncertainty when evidence is insufficient or conflicting.
These constraints are enforced through inference and self-checking prompts.

\textbf{Reconstruction prompt templates.}
The following condensed prompt excerpts specify the three
LLM inference stages and their programmatic input/output
contracts.

\begingroup

\definecolor{snoopnavy}{HTML}{2B4053}
\definecolor{snooppaper}{HTML}{F6F7F9}
\definecolor{snoopred}{HTML}{B52427}

\tcbset{
  recoveryprompt/.style={
    enhanced,
    colframe=snoopnavy,
    colback=snooppaper,
    colbacktitle=snoopnavy,
    coltitle=white,
    fonttitle=\scriptsize\bfseries,
    fontupper=\scriptsize,
    boxrule=0.5pt,
    arc=1.5pt,
    left=5pt,
    right=5pt,
    top=4pt,
    bottom=4pt,
    toptitle=3pt,
    bottomtitle=3pt,
    before skip=5pt,
    after skip=5pt,
    before upper={
      \raggedright
      \setlength{\parindent}{0pt}
      \setlength{\parskip}{1pt}
    }
  }
}

\begin{tcolorbox}[
  recoveryprompt,
  title={Prompt 1: Evidence-Grounded Intent Inference},
  top=2pt, bottom=2pt,
  left=4pt, right=4pt,
  toptitle=2pt, bottomtitle=2pt,
  before upper={
    \raggedright
    \setlength{\parindent}{0pt}
    \setlength{\parskip}{0pt}
  }
]
You are given confirmed Web-selection paths from an agent
executing an \emph{unknown user query}. Infer what the
user intended to accomplish, not merely the topic of
the pages visited.

\textbf{Input:}
\texttt{[confirmed\_paths]},
\texttt{[node\_descriptions]},
\texttt{[path\_ids]}.\\
\textbf{Inference interface:}
\texttt{INFERINTENT}($\mathcal{P}_q$)
$\rightarrow$ \texttt{IntentRecord}.

\textcolor{snoopred}{\textbf{IMPORTANT:}
Use only confirmed, selected nodes; unselected
candidates and missing requests are not evidence.}

\begin{enumerate}[
  leftmargin=1.3em,
  label=\arabic*.,
  topsep=0pt,
  itemsep=0pt,
  parsep=0pt,
  partopsep=0pt
]
\item Read each ordered path and infer its subtask-level
information need; retain supporting node references.
\item Classify supported intents as \emph{identify,
explain, compare, assess, recommend}, or \emph{quantify}.
\item Consolidate subtask intents into the primary query
intent; retain a secondary intent only if supported.
\item Check intent--evidence consistency and mark
ambiguous or insufficiently supported predictions.
\end{enumerate}
\noindent\emph{...}\\
\textbf{Output:}
\texttt{primary\_intent, secondary\_intent,
subtask\_intents[...], evidence[...], status}.
\end{tcolorbox}

\subsection{Scoring Metrics}
\label{app:metrics}

\textbf{Query-level evaluation.}
We measure primary-intent accuracy by exact match and
concept recovery using Precision, Recall, and F1.
Let $P=\{p_1,\ldots,p_m\}$ and $G=\{g_1,\ldots,g_n\}$
denote the recovered and ground-truth Wikidata concepts.
Their pairwise matching score is

{\scriptsize
\begin{equation}
s(p_i,g_j)=
\begin{cases}
1, & p_i=g_j,\\
0.5, & 1\leq d_{\mathrm{hier}}(p_i,g_j)\leq 2,\\
0, & \text{otherwise},
\end{cases}
\end{equation}
}

\begin{tcolorbox}[
  recoveryprompt,
  title={Prompt 2: Multi-Path Query Reconstruction},
  top=2pt, bottom=2pt,
  left=4pt, right=4pt,
  toptitle=2pt, bottomtitle=2pt,
  before upper={
    \raggedright
    \setlength{\parindent}{0pt}
    \setlength{\parskip}{0pt}
  }
]
Reconstruct \emph{one coherent natural-language query}
from the inferred intent and confirmed subtask paths.
Recover the information need, not the original wording.

\textbf{Input:}
\texttt{[recovered\_intent]},
\texttt{[confirmed\_paths]},
\texttt{[node\_descriptions]}.\\

\textbf{Inference interface:}
\texttt{RECONSTRUCTQUERYTEXT}($\mathcal{P}_q,I$)
$\rightarrow$ \texttt{QueryRecord}.

\textcolor{snoopred}{\textbf{IMPORTANT:}
Every substantive entity, relation, comparison,
or constraint must have a supporting path reference.
Do not complete missing details from prior knowledge.}

\begin{enumerate}[
  leftmargin=1.3em,
  label=\arabic*.,
  topsep=0pt,
  itemsep=0pt,
  parsep=0pt,
  partopsep=0pt
]
\item Extract supported entities, attributes, relations,
and constraints; track their source paths.
\item Combine complementary subtask evidence, resolving
overlaps without losing distinct information.
\item Form a query aligned with the primary intent,
preserving supported comparisons and conditions.
\item Audit the query against confirmed paths; remove
unsupported details and retain uncertainty.
\end{enumerate}
\noindent\emph{...}\\
\textbf{Output:}
\texttt{reconstructed\_query, primary\_intent,
evidence[...], status}.
\end{tcolorbox}

\begin{tcolorbox}[
  recoveryprompt,
  title={Prompt 3: Cross-Query Profile Reconstruction},
  top=2pt, bottom=2pt,
  left=4pt, right=4pt,
  toptitle=2pt, bottomtitle=2pt,
  before upper={
    \raggedright
    \setlength{\parindent}{0pt}
    \setlength{\parskip}{0pt}
  }
]
You are given evidence-linked query reconstructions
from successive interaction rounds. Infer a categorized
user profile from their combined semantic evidence,
without access to the original conversation or PTKB.

\textbf{Input:}
\texttt{[query\_representations]}
with $Q_i=\langle I_i,E_i,R_i\rangle$;
\texttt{[query\_ids], [evidence\_refs]}.\\
\textbf{Inference interface:}
\texttt{AGGREGATEPROFILE}($\mathcal{Q}$)
$\rightarrow$ \texttt{ProfileRecord}.

\textcolor{snoopred}{\textbf{IMPORTANT:}
A researched topic is not itself a personal attribute.
Attribute facts to the user only when supported by
reconstructed query context.}

\begin{enumerate}[
  leftmargin=1.3em,
  label=\arabic*.,
  topsep=0pt,
  itemsep=0pt,
  parsep=0pt,
  partopsep=0pt
]
\item Align related Wikidata concepts and query evidence
across rounds; identify corroboration and conflicts.
\item Infer supported attributes, interests, and preferences
within appropriate categories.
\item Merge equivalent facts and attach supporting query IDs
to every retained claim.
\item Mark conflicting or weakly supported facts as uncertain;
exclude unsupported claims from the confirmed profile.
\end{enumerate}
\noindent\emph{...}\\
\textbf{Output:}
\texttt{profile[category, facts(fact, evidence, status),
...]; conflicts[...]}.
\end{tcolorbox}

\endgroup

where $d_{\mathrm{hier}}$ is the ancestor--descendant distance
along Wikidata's \texttt{P279} (\emph{subclass of}) and
\texttt{P31} (\emph{instance of}) relations.
Sibling and unrelated concepts score zero.
Precision and Recall are computed as
$\mathrm{Prec}=\tfrac{1}{|P|}\sum\nolimits_i
\max\nolimits_j s(p_i,g_j)$ and
$\mathrm{Rec}=\tfrac{1}{|G|}\sum\nolimits_j
\max\nolimits_i s(p_i,g_j)$,
respectively, with F1 as their harmonic mean.
Table~\ref{tab:scoring-example} illustrates the matching:
four pairs share identical Wikidata Q-IDs (including
\emph{blood glucose} and \emph{blood sugar}), while
\emph{endocrine disease} partially matches
\emph{type 2 diabetes} within two hops.
This yields $\mathrm{Prec}=\mathrm{Rec}=\mathrm{F1}=0.90$.

\textbf{Profile-level evaluation.}
We compare recovered profile facts against the original
iKAT PTKB using the same Precision/Recall/F1 formulas.
An LLM judge assigns each predicted--ground-truth pair
a score of $1$ (equivalent), $0.5$ (partial match), or
$0$ (unrelated). Only facts within the same attribute
category can match, and all judge scores undergo manual
verification. In Table~\ref{tab:profile-example},
three exact matches, one partial match
(\emph{technical degree}/\emph{computer science}),
and one unmatched fact (\emph{acid reflux}) yield $\mathrm{F1}\approx0.78$.

\begin{table}[t]
\centering
\caption{Pairwise Wikidata concept matching scores.}
\label{tab:scoring-example}
\scriptsize
\setlength{\tabcolsep}{6pt}
\renewcommand{\arraystretch}{1}
\begin{tabular}{@{}lccccc@{}}
\toprule
\textbf{$s(p,g)$} &
\makecell{type 2\\diabetes} &
diet &
\makecell{physical\\exercise} &
\makecell{insulin\\resistance} &
\makecell{blood\\sugar} \\
\midrule
endocrine disease  & 0.5 & 0 & 0 & 0 & 0 \\
diet               & 0 & 1 & 0 & 0 & 0 \\
physical exercise  & 0 & 0 & 1 & 0 & 0 \\
insulin resistance & 0 & 0 & 0 & 1 & 0 \\
blood glucose      & 0 & 0 & 0 & 0 & 1 \\
\bottomrule
\end{tabular}
\end{table}

\begin{table}[t]
\centering
\caption{Pairwise profile fact scores across
attributes.}
\label{tab:profile-example}
\scriptsize
\setlength{\tabcolsep}{1pt}
\renewcommand{\arraystretch}{1}
\begin{tabular}{@{}lcccc@{}}
\toprule
\multirow{2}{*}{\textbf{Recovered fact}} &
\multicolumn{4}{c}{\textbf{Ground-truth fact}} \\
\cmidrule(lr){2-5}
& \makecell{\emph{Nether-}\\\emph{lands} (loc.)}
& \makecell{\emph{software}\\\emph{engineer} (occ.)}
& \makecell{\emph{computer}\\\emph{science} (edu.)}
& \makecell{\emph{cycling}\\(int.)} \\
\midrule
\emph{Netherlands} (loc.)        & 1 & 0 & 0 & 0 \\
\emph{acid reflux} (health)      & 0 & 0 & 0 & 0 \\
\emph{software developer} (occ.) & 0 & 1 & 0 & 0 \\
\emph{technical degree} (edu.)   & 0 & 0 & 0.5 & 0 \\
\emph{cycling} (int.)            & 0 & 0 & 0 & 1 \\
\bottomrule
\end{tabular}
\end{table}

\textbf{Induction rate.}
We define $\mathrm{IR}=|\{t\in\mathcal{T}:\mathrm{path}(t)\neq\emptyset\}|/|\mathcal{T}|$,
where $\mathcal{T}$ includes all attack trials.
Trials without exposure to an injected page, link following,
or a decodable path count as failures.

\textbf{Baseline scoring.}
We evaluate each baseline using its native recovery
target and metric, with ground truth derived from
frozen benchmark annotations or predefined attack targets.
(i) \emph{WRA} predicts privacy groups from domain sequences.
On Stack, predicted groups are matched against annotated
domain labels; on iKAT, recovered traits are matched
against PTKB-derived trait groups.
We compute group-level Precision and Recall from the
overlap between predicted and ground-truth label sets,
and report their harmonic-mean F1.
(ii) \emph{AgentPrint} classifies encrypted traffic into coarse
privacy categories derived from benchmark annotations.
We report top-1 accuracy as the fraction of traces whose
highest-ranked prediction matches the ground-truth
category, and top-3 accuracy when the correct category
appears among the three highest-ranked predictions.
(iii) \emph{SkillWatermark} encodes predefined attribute values
into conditional traffic signatures. We measure trigger
rate over all trials and conditional decoding accuracy
over triggered trials by comparing decoded values with
the encoded ground truth, which is known by construction.
(iv) \emph{Explicit exfiltration} targets designated private
data in the agent's task context. We measure success rate
as the fraction of trials in which the complete target
secret reaches an attacker-controlled endpoint verbatim;
partial or paraphrased disclosures count as failures.
Ground truth is used only for post-execution verification.
These metrics reflect different recovery targets.
We therefore compare both recovery granularity and
effectiveness under each method's native scoring criterion.

\section{Attack and Defense Baseline Implementations}
\label{app:baselines}

All baselines share the same configurations while retaining their native observation capabilities: WRA observes domain-access logs, AgentPrint captures encrypted traffic metadata, and endpoint-based attacks monitor only their own requests. Private prompts, agent reasoning, and ground truth are withheld during inference.

\subsection{Attack Baselines}

Table~\ref{tab:baseline-repro} summarizes the observation
and training requirements of each attack.

\begin{table}[t]
\centering
\caption{Attack baseline configurations.}
\label{tab:baseline-repro}
\scriptsize
\setlength{\tabcolsep}{10pt}
\renewcommand{\arraystretch}{1}
\begin{tabular}{@{}lll@{}}
\toprule
\textbf{Method} &
\textbf{Observation} &
\textbf{Training / Calibration} \\
\midrule
WRA~\cite{jeong2026wra}
& Domain sequences & 5-shot ICL \\
AgentPrint~\cite{zhang2025agentprint}
& Encrypted traffic & CNN (80/20 split) \\
SkillWatermark~\cite{li2026skillwatermark}
& Traffic size/timing & Decoder (200 traces) \\
Explicit exfil.~\cite{zhan2024injecagent}
& Attacker endpoint & None \\
\midrule
\rowcolor{snoopblue!8}
\textbf{\sys (ours)}
& Attacker endpoint & None \\
\bottomrule
\end{tabular}
\end{table}

\textbf{WRA.}
We serialize visited-domain logs into ordered
(domain, timestamp, bytes) records and filter
advertising/tracking hosts using
EasyList~\cite{easylist} and
EasyPrivacy~\cite{easyprivacy}, reducing sequence
length from 58 to 40 entries per task (30\%).
Following WRA~\cite{jeong2026wra}, we use 5-shot ICL.
The demonstrations come from 60 tasks and three personas
disjoint from evaluation. Stack predictions are mapped
to domain groups; iKAT combines successive turns
for persona-level trait inference.

\textbf{AgentPrint.}
We transform encrypted traffic (packet direction, size,
and timing) into $2\times2\times1800$ Multi-view Traffic
Aggregation Matrices (MTAMs), encoding direction,
packet count, byte volume, and temporal windows.
Following AgentPrint~\cite{zhang2025agentprint},
the classifier uses two Conv2D blocks (32/64 filters,
$3\times3$ kernels), two Conv1D blocks (128 filters),
global average pooling, and a dataset-specific softmax head.
Training uses Adam ($10^{-3}$), cross-entropy,
batch size 64, and 60 epochs.
The 10,536 traces are split 80/20, disjoint by query on Stack and by
persona on iKAT; all traces of the same underlying
task remain in one partition.

\textbf{SkillWatermark.}
We adapt it by moving conditional directives from
\texttt{Skill.md} to mirrored Web content while retaining
traffic-based encoding and decoding.
A 12-term codebook covers response-length and
resource-retrieval behaviors, with signatures calibrated
using packet-size differences $\geq1.8$\,KB or
inter-packet gaps $\geq120$\,ms.
Up to four bits are encoded per session across turns.
A decoding dictionary is calibrated from 200 held-out
triggered traces. We report trigger rate and conditional
decoding accuracy.

\textbf{Explicit exfiltration.}
We adapt it
by injecting its data-stealing templates into mirrored
pages, using URL query parameters, tool arguments,
and Markdown image requests as exfiltration carriers.
Both base and enhanced attack variants are evaluated.
Success requires the complete secret to reach the
attacker endpoint verbatim; partial or paraphrased
leaks count as failures. The 175 non-completing runs
among 4,216 trials are also counted as failures.

\subsection{Defense Baselines}
\label{app:defenses}

Table~\ref{tab:defense-repro} summarizes the four
defenses and their deployment configurations.

\begin{table}[t]
\centering
\caption{Defense baseline configurations.}
\label{tab:defense-repro}
\scriptsize
\setlength{\tabcolsep}{11pt}
\renewcommand{\arraystretch}{1}
\begin{tabular}{@{}lll@{}}
\toprule
\textbf{Defense} &
\textbf{Enforcement layer} &
\textbf{Configuration} \\
\midrule
Spotlighting~\cite{hines2024spotlighting}
& Prompt/input & Datamarking (\texttt{U+E000}) \\
Firewalls~\cite{sharma2025firewalls}
& Tool-I/O & Sanitizer + Minimizer \\
Task Shield~\cite{jia2025taskshield}
& Action & Task-alignment threshold 0.5 \\
Fides~\cite{costa2025fides}
& Information flow & $2\times2$ security-label lattice \\
\bottomrule
\end{tabular}
\end{table}

\textbf{Spotlighting.}
We apply datamarking~\cite{hines2024spotlighting}
to all fetched pages and search snippets, replacing
inter-word spaces with \texttt{U+E000}.
A system-level instruction marks these spans as
untrusted data before they enter agent reasoning.
No additional detection model is used.

\textbf{Firewalls.}
We deploy Firewalls's~\cite{sharma2025firewalls} Sanitizer and Minimizer
at the
Web search/fetch boundary.
The Sanitizer removes suspected injections from
incoming tool results, while the Minimizer filters
outgoing search queries and fetch URLs to restrict
unnecessary information disclosure.

\textbf{Task Shield.}
We apply Task Shield~\cite{jia2025taskshield}
before each Web search or fetch call.
A task-alignment check evaluates each call's contribution
to the user task and permits actions with a
\textsc{ContributeTo} score $\geq0.5$ on $[0,1]$.

\textbf{Fides.}
We implement Fides~\cite{costa2025fides}
using a $2\times2$ confidentiality
(public/confidential) and integrity
(trusted/untrusted) lattice.
User queries and private context are labeled
confidential, while retrieved content is
untrusted. Outbound tool actions are checked against these labels before execution.

\section{Behavioral Channel Encoding and Decoding}
\label{app:channels}

All six channels infer the agent's current subtask among candidate $S_1$--$S_4$, differing in observable signals and request costs. The attacker predefines and rotates codeword mappings across trials, decoding solely from requests to attacker-controlled endpoints.

\textbf{Choice.} The page presents $k$ topic-specific links $A$--$D$, one per candidate. Pursuing its need, the agent requests the matching link, and the requested endpoint identifies the candidate in a single request.

\textbf{Binary.} Each candidate is a $\lceil\log_2 k\rceil$-bit word ($S_1$--$S_4 = 00,01,10,11$). Each round exposes one optional beacon for the current bit; a request decodes $1$ and its absence decodes $0$. Two rounds resolve $k=4$.

\textbf{Count.} The candidates form a sequential catalog behind a \texttt{/next} action and a shared \texttt{/confirm}. The agent issues \texttt{/next} to advance to its candidate at position $i$, then \texttt{/confirm}; the number of \texttt{/next} requests decodes $i$.

\textbf{Sequence.} The page exposes $m$ shared resources whose relevance order depends on the candidate ($m!\ge k$, so $m=3$ for $k=4$). The agent fetches them in its relevance order, and the observed request order decodes the candidate.

\textbf{Timing.} Candidates are assigned a delay bin using staggered resource-ready times behind a shared \texttt{/continue} endpoint. The agent requests \texttt{/continue} when ready, and the elapsed time from page load identifies the bin.

\textbf{Size.} Each candidate is mapped to a byte-size bin of a \texttt{/report} section. The agent issues a Range request for the section it needs, and the transferred byte count decodes the bin. Whole-file fetching transfers a constant volume and does not support this channel.

\end{document}